\documentclass{article}
\usepackage[T1]{fontenc}

\usepackage{microtype}
\usepackage{graphicx}
\usepackage{subcaption}
\usepackage{booktabs}
\usepackage{array}
\usepackage{multirow}
\usepackage[table]{xcolor}
\usepackage{hyperref}

\usepackage[preprint]{icml2026}
\usepackage{algorithm}
\usepackage{algorithmic}

\usepackage{amsmath}
\usepackage{amssymb}
\usepackage{mathtools}
\usepackage{amsthm}
\usepackage{bm}
\usepackage{enumitem}
\usepackage[capitalize,noabbrev]{cleveref}

\theoremstyle{plain}
\newtheorem{theorem}{Theorem}
\newtheorem{proposition}[theorem]{Proposition}

\newtheorem{corollary}[theorem]{Corollary}
\theoremstyle{definition}
\newtheorem{definition}{Definition}
\theoremstyle{remark}

\newcommand{\tg}{Team-GRPO}
\newcommand{\E}{\mathbb{E}}

\newcommand{\own}{\mathrm{own}}
\newcommand{\ext}{\mathrm{ext}}
\newcommand{\sth}{s_\theta}
\newcommand{\pth}{\pi_\theta}
\newcommand{\Proj}{\mathrm{Proj}_\Delta}
\newcommand{\rbar}{\bar r}
\newcommand{\ind}{\mathbf{1}}
\newcommand{\gr}{\mathcal{G}}

\let\oldtextsc\textsc
\renewcommand{\textsc}[1]{\textup{\oldtextsc{#1}}}


\definecolor{cteam}{HTML}{2A78D6}
\definecolor{cgrpo}{HTML}{EB6834}
\definecolor{cnash}{HTML}{1BAF7A}

\icmltitlerunning{Group-Dependent Rewards and GRPO}

\begin{document}

\twocolumn[
  \icmltitle{When Rewards Depend on the Group, GRPO Plays a Tournament}

  \begin{center}
    {\large
      \textbf{Junyi Gao}\textsuperscript{1,3,*}\quad
      \textbf{Yu Shi}\textsuperscript{2,*}\quad
      \textbf{Pingzhao Hu}\textsuperscript{4,\textdagger}\quad
      \textbf{Ewen M Harrison}\textsuperscript{1,\textdagger}\par}
    \vspace{0.5em}
    \textsuperscript{1}Centre for Medical Informatics, University of Edinburgh\\
    \textsuperscript{2}Biostatistics Division, Dalla Lana School of Public Health, University of Toronto\\
    \textsuperscript{3}Health Data Research UK\\
    \textsuperscript{4}Department of Biochemistry, Western University\\
    \textsuperscript{*}Equal contribution\qquad
    \textsuperscript{\textdagger}Senior author
  \end{center}
  \hypersetup{pdfauthor={Junyi Gao, Yu Shi, Pingzhao Hu, Ewen M Harrison}}
  \icmlkeywords{reinforcement learning, policy gradient, GRPO, credit assignment, diversity, language models}

  \vskip 0.3in
]

\makeatletter
\renewcommand{\printAffiliationsAndNotice}[1]{%
  \global\icml@noticeprintedtrue
  {\let\thefootnote\relax
   \footnotetext{\hspace*{-\footnotesep}\Notice@String}}%
}
\makeatother
\printAffiliationsAndNotice{}

\begin{abstract}
Rewards for diversity, coverage and distribution matching make the reward of each sampled response depend on the rest of its group, and they are now routinely optimized with GRPO and RLOO. We prove that in this setting the group-relative baseline changes the objective. The expected update equals each response's own reward gradient plus $\alpha$ times its effect on its groupmates' rewards, with $\alpha=-1/(K-1)$ for groups of size $K$, so GRPO trains every response to beat its groupmates in a zero-sum tournament. It cannot see a pairwise diversity bonus when $K=2$, sees less than half of it at any $K$, and often does worse than using no baseline at all. Setting $\alpha=+1$ gives \tg{}, an unbiased estimator for any group-dependent reward that works with two samples per prompt and requires no extra model calls for pairwise rewards. The theory predicts training quantitatively. It locates the fixed points of a 9B policy to within $0.02$ in $\ell_1$ and predicts how concentrated trained policies become with Pearson $r=0.995$. On GlobalOpinionQA, GRPO collapses onto the majority answer and ends further from human opinion than the untrained model, while \tg{} cuts the error from $0.20$ to $0.08$. On NoveltyBench, \tg{} avoids the diversity collapse of GRPO and significantly improves set utility.
\end{abstract}

\section{Introduction}\label{sec:intro}

Group-relative policy gradients are the default way to fine-tune language models with reinforcement learning. GRPO \citep{shao2024deepseekmath} and RLOO \citep{kool2019buy,ahmadian2024back} sample $K$ responses per prompt and score each against the mean reward of its group, which removes the need for a learned critic \citep{guo2025deepseek,liu2025understanding}. The justification is classical \citep{williams1992simple,greensmith2004variance}. When the reward of a response depends only on that response, its groupmates' rewards are independent of it, so subtracting their mean lowers variance without changing the expected update.

A fast-growing family of rewards breaks this assumption. To counter the loss of diversity under reinforcement learning \citep{kirk2024understanding,yue2025does}, DARLING multiplies each response's quality by its semantic distance to its groupmates \citep{li2025jointly}, and other methods add pairwise novelty bonuses \citep{song2025outcome,dsdr2026} or reward answers that are rare within the group \citep{anschel2025gapo}. Distributional alignment penalizes collisions with groupmates so that the group reproduces a target distribution \citep{swys2026}, and set-level objectives score the group as a whole \citep{walder2025pkpo,orney2026polyepo}. In each case the reward of response $k$ depends on the whole group, so its baseline depends on it through its groupmates' rewards, and the classical argument fails.

\begin{figure}[t]
  \centering
  \includegraphics[width=0.84\columnwidth]{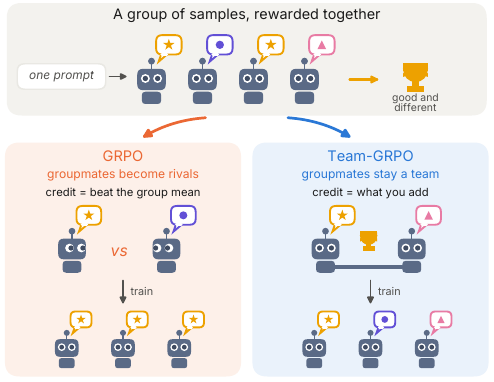}
  \vspace{-0.9em}
  \caption{\textbf{The problem.} An answer helps its group by being good and by being different. GRPO credits each sample with how far it beats the group mean, so groupmates become rivals and converge on one answer. \tg{} credits each sample with what it adds to the group, so the group stays diverse.}
  \label{fig:insight}
  \vspace{-1.0em}
\end{figure}

We show that the result is not extra noise but a different objective. For any reward that treats groupmates symmetrically, the expected GRPO update is the gradient of a response's own reward plus $\alpha$ times its \emph{externality}, the change it causes in its groupmates' rewards, with $\alpha=-1/(K-1)$ (\cref{thm:identity}). The true gradient counts the externality with weight $+1$, so subtracting the group mean does not merely discard it but reverses its sign. GRPO trains each response to beat its groupmates, a zero-sum tournament like the relative-payoff game that governs selection in finite populations \citep{schaffer1988evolutionarily}, in which groupmates are rivals rather than a team (\cref{fig:insight}).

\begin{figure*}[t]
  \centering
  \includegraphics[width=0.94\textwidth]{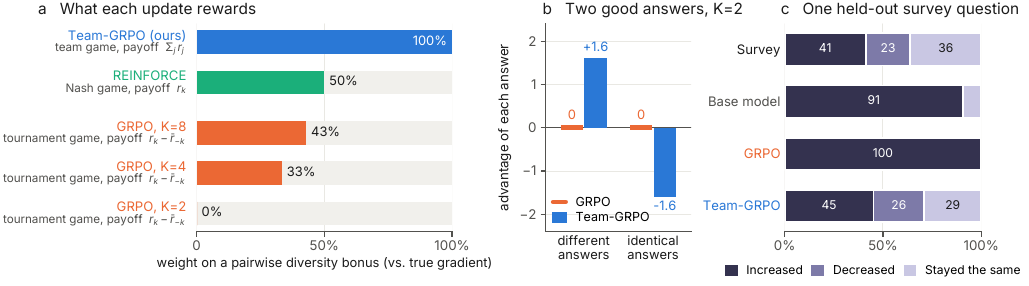}
  \vspace{-0.6em}
  \caption{\textbf{GRPO plays a tournament.}
  \textbf{(a)}~\tg{} maximizes the group's total reward, REINFORCE without a baseline maximizes each response's own reward (a Nash equilibrium), and GRPO maximizes each response's reward minus the mean of its groupmates' (a zero-sum tournament). Bars give the weight of a pairwise diversity bonus in the expected update relative to the true gradient (\cref{cor:ladder}).
  \textbf{(b)}~Two responses of quality $q=0.8$ under the reward $q\,(1+\lambda d)$, $\lambda=2$. GRPO assigns zero advantage whether the answers differ ($d=1$) or coincide ($d=0$). \tg{} separates the two cases.
  \textbf{(c)}~Answer shares on a held-out GlobalOpinionQA question (whether the gap between rich and poor has grown, Venezuela) after 200 steps with two samples per question.}
  \label{fig:concept}
  \vspace{-0.9em}
\end{figure*}

The tournament has closed-form consequences (\cref{fig:concept}). A pairwise diversity bonus reaches the update with $(K-2)/(2(K-1))$ of its true weight, which is below one half for every $K$ and zero at $K=2$, where GRPO cannot tell two different good answers from one good answer given twice. Larger groups do not rescue it, because as $K$ grows GRPO approaches REINFORCE without a baseline rather than the optimum of the reward. Across six reward structures from recent methods, GRPO is never better than REINFORCE and is often far worse.

Setting $\alpha=+1$ restores the true gradient. The resulting estimator, \tg{}, credits each response with its marginal contribution to the group's reward, using only quantities the reward already computes (\cref{sec:method}). It works with two samples per prompt, an efficient regime \citep{twogrpo2025} in which GRPO fails completely, and it contains recent special-purpose estimators for max@$k$ \citep{walder2025pkpo,takashiro2026l2o}, distribution matching \citep{swys2026} and set-level reinforcement learning \citep{orney2026polyepo} as special cases.

The theory makes quantitative predictions, each fixed before the corresponding experiment. In a sandbox where the policy of a 9B model can be read off exactly, all 12 runs with a closed-form prediction converge to the fixed point of their estimator's game (largest $\ell_1$ error $0.018$). On GlobalOpinionQA \citep{durmus2024towards}, the theory predicted that GRPO with two samples per question would collapse onto the majority answer and end worse than the untrained model. It did, raising the squared error to the human distribution from $0.20$ to $0.49$, while \tg{} lowers it to $0.08$. Across 13 configurations the closed-form fixed points predict how concentrated the trained policy becomes with Pearson $r=0.995$. On NoveltyBench \citep{zhang2025noveltybench} with a quality-times-diversity reward, GRPO collapses in every seed, whereas \tg{} keeps the model diverse and raises set utility over GRPO by $1.16$ at $K=2$ and $0.56$ at $K=8$, gains that persist at matched KL divergence and under an independent LLM judge.

\paragraph{Contributions.}
\begin{itemize}[leftmargin=1.2em,itemsep=1pt,topsep=2pt]
  \item \textbf{Diagnosis.} An exact identity for group-relative estimators under any exchangeable group reward, its reading as a relative-payoff game, and its closed-form consequences (\cref{sec:theory}).
  \item \textbf{Method.} \tg{}, an unbiased and essentially free estimator for any exchangeable group reward that works at $K=2$ and unifies recent special-purpose estimators (\cref{sec:method}).
  \item \textbf{Evidence.} Quantitative confirmation on a 9B model in distributional alignment and open-ended generation, where \tg{} outperforms GRPO and its variants and matches set-level estimators wherever they apply (\cref{sec:exp}).
\end{itemize}

\section{What Group-Relative Baselines Optimize}\label{sec:theory}

\paragraph{Setup.}
Fix a prompt, which we drop from the notation. The policy $\pth$ samples a group $\gr=(y_1,\dots,y_K)$ of independent responses, and response $k$ receives a reward $r_k$ that may depend on the whole group but treats the groupmates symmetrically.

\begin{definition}[Exchangeable group reward]\label{def:exch}
The reward is \emph{exchangeable} if $r_k=r\big(y_k;\{y_j\}_{j\neq k}\big)$ for a fixed function $r$ of a response and the multiset of its groupmates. The group objective is $J(\theta)=\E\big[\tfrac1K\sum_{k}r_k\big]$.
\end{definition}

Ordinary rewards $r_k=f(y_k)$ are the special case in which $J$ is the expected reward, and every reward in \cref{sec:intro} is exchangeable (\cref{app:methods}). Write $\sth(y)=\nabla_\theta\log\pth(y)$. A group-relative estimator assigns an advantage $A_k$ to each response and uses $\hat g=\sum_k A_k\,\sth(y_k)$. GRPO uses $A_k=r_k-\rbar$ with $\rbar=\tfrac1K\sum_j r_j$, the mean-centred form adopted by Dr.\ GRPO and DARLING \citep{liu2025understanding,li2025jointly}, RLOO uses $A_k=r_k-\tfrac{1}{K-1}\sum_{j\neq k}r_j$, and REINFORCE uses a baseline that does not depend on the group (standard-deviation normalization is treated in \cref{app:std}). Two functions of a single response organize everything that follows, its \emph{own value} and its \emph{externality},
\begin{equation}\label{eq:ownext}
  \own(y)=\E\big[r_1\mid y_1=y\big],\;\; \ext(y)=\sum_{j=2}^{K}\E\big[r_j\mid y_1=y\big].
\end{equation}
The first is the reward a response can expect for itself, and the second is the total reward its groupmates can expect, which is constant for ordinary rewards.

\subsection{An exact identity}

\begin{theorem}[The $\alpha$-identity]\label{thm:identity}
For any exchangeable group reward, let $g_\alpha=\E_{y\sim\pth}\big[\big(\own(y)+\alpha\,\ext(y)\big)\,\sth(y)\big]$. Then
\begin{gather*}
  \nabla_\theta J=g_{1},\qquad \E\,\hat g_{\textsc{reinforce}}=K\,g_{0},\\
  \E\,\hat g_{\textsc{rloo}}=K\,g_{-\frac{1}{K-1}},\qquad \E\,\hat g_{\textsc{grpo}}=(K-1)\,g_{-\frac{1}{K-1}}.
\end{gather*}
\end{theorem}

\begin{proof}[Proof sketch]
For $j\neq k$, conditioning on $y_k$ gives $\E[r_j\,\sth(y_k)]=\E\big[\E[r_j\mid y_k]\,\sth(y_k)\big]$, so the RLOO baseline contributes $-\tfrac{1}{K-1}\ext$ and a group-independent baseline nothing. The true gradient follows from $\nabla_\theta\E\big[\sum_j r_j\big]=\sum_k\E\big[\big(\sum_j r_j\big)\sth(y_k)\big]=Kg_1$ (\cref{app:proofs}).
\end{proof}

The three estimators thus lie on one line indexed by $\alpha$, the weight the update gives to externalities, which we call the $\alpha$-line. Positive multiples only rescale the step, so the estimators differ in direction through $\alpha$ alone. The true gradient counts a response's effect on others with weight $+1$, REINFORCE ignores it, and GRPO counts it with weight $-1/(K-1)$, penalizing a response that raises its groupmates' rewards and rewarding one that lowers them.

\paragraph{Which game is being played.}
Give player $k$ of a symmetric $K$-player game the payoff $U^\alpha_k=r_k+\alpha\sum_{j\neq k}r_j$. Then $g_\alpha$ is the gradient of player $k$'s expected payoff with respect to its own policy when all players use $\pth$, so training with an $\alpha$-estimator is self-play with shared parameters and its stationary points are symmetric equilibria of this game (\cref{prop:game}). For $\alpha=1$ all players share the payoff $\sum_j r_j$, a team game whose optimum maximizes $J$, and for $\alpha=0$ each response maximizes its own reward, a Nash equilibrium. For GRPO, $U_k=r_k-\rbar_{-k}$ is a response's reward relative to the mean of its groupmates. These payoffs sum to zero, so GRPO plays a zero-sum tournament, the relative-payoff selection of finite populations that favors spite \citep{schaffer1988evolutionarily,hamilton1970selfish}, with the group as the population. As $K\to\infty$ its $\alpha$ tends to $0$, so larger groups move GRPO toward the Nash solution but never toward the team optimum.

\paragraph{Which weight is intended.}
The diagnosis does not rest on taking $J$ as the target. A designer may want each response credited for its total effect on the group ($\alpha=1$) or for its own reward against fixed groupmates ($\alpha=0$), but never for lowering its groupmates' rewards. GRPO's $\alpha$ is negative at every $K$, so it serves neither intention, and at $K=2$ it gives a pairwise diversity bonus zero weight where both give it positive weight. The methods we study intend $\alpha=1$, since each is evaluated by a property of the whole set, such as pass@$k$, coverage or distance to a target distribution (\cref{tab:methods}). For $\alpha=0$ the same construction gives an unbiased estimator (\cref{app:family}).

\subsection{Consequences in closed form}

\begin{corollary}[Attenuation ladder]\label{cor:ladder}
Let $r_k=f(y_k)+\lambda\,c_k$, where $c_k$ averages a symmetric kernel $\phi(y_k,y_{S})$ over all sets $S$ of $m-1$ groupmates ($m=2$ for pairwise diversity, similarity and collision terms). With $\Phi(y)=\E[\phi(y,Y_2,\dots,Y_m)]$,
\begin{equation}\label{eq:ladder}
  g_\alpha=\E\big[\big(f+\lambda\,(1+\alpha(m-1))\,\Phi\big)\sth\big].
\end{equation}
The true gradient weights the interaction by $m$ and GRPO by $\tfrac{K-m}{K-1}$, so GRPO retains a fraction
\begin{equation}\label{eq:retain}
  \rho_m(K)=\frac{K-m}{m(K-1)}<\frac1m .
\end{equation}
If $\sum_k c_k$ is constant, as for within-group win rates, the weight is $1-\alpha$ and GRPO amplifies the term by $\tfrac{K}{K-1}$ although it cannot change $J$. If the reward is shared by the whole group, as for pass@$k$ credited to every member, the weight is $1+\alpha(K-1)$ and GRPO sees none of it.
\end{corollary}

A pairwise bonus thus keeps $0$, $\tfrac13$ and $\tfrac37$ of its weight at $K=2,4,8$ (\cref{fig:concept}a), never more than one half, and three-body terms keep less than a third.

\begin{proposition}[Pairwise rewards]\label{prop:pairwise}
Let $r_k=f(y_k)+\tfrac{\lambda}{K-1}\sum_{j\neq k}M(y_k,y_j)$ for an arbitrary kernel $M=S+A$ with symmetric part $S$ and antisymmetric part $A$. With $\bar S(y)=\E_{y'}S(y,y')$ and $\bar A(y)=\E_{y'}A(y,y')$,
\begin{equation}\label{eq:pairwise}
  g_\alpha=\E\big[\big(f+\lambda[(1+\alpha)\bar S+(1-\alpha)\bar A]\big)\sth\big].
\end{equation}
\end{proposition}

The true gradient contains only $\bar S$, since an antisymmetric interaction moves reward between responses without changing the total. GRPO at $K=2$ ($\alpha=-1$) contains only $\bar A$, doubled, so \emph{every} symmetric interaction, including any diversity bonus, is invisible to it. For $K>2$, GRPO mixes in $\bar A$ with weight $K/(K-1)$. This part is a rotational field, so the GRPO update is not the gradient of any objective (\cref{app:proofs}).

\paragraph{Example.} DARLING-style rewards~\citep{li2025jointly} $r_k=q(y_k)(1+\lambda D_k)$, with $D_k$ the mean distance to the groupmates, have $M(y,y')=q(y)\,d(y,y')$. At $K=2$ the GRPO advantage is
\begin{equation}\label{eq:k2adv}
  A_1^{\textsc{grpo}}=\tfrac12\,(q_1-q_2)\,(1+\lambda d_{12}).
\end{equation}
It ranks the two responses by quality and uses their distance only to scale the ranking, so two different good answers and two copies of one good answer both receive zero advantage (\cref{fig:concept}b). Raising $\lambda$ even hurts, because it amplifies the quality ranking in exactly the most diverse pairs.

\begin{proposition}[Fixed points]\label{prop:fixed}
On a finite answer space with a softmax policy, the expected $\alpha$-update ascends an explicit potential in two canonical cases.
\begin{enumerate}[label=(\alph*),leftmargin=1.4em,itemsep=1pt,topsep=2pt]
  \item \emph{Diversity bonus.} For $r_k=f(y_k)+\tfrac{\lambda}{K-1}\sum_{j\neq k}\ind[y_k\neq y_j]$ the potential is $\E_\pi f+\kappa_\alpha\lambda(1-\|\pi\|^2)$ with $\kappa_\alpha=\tfrac{1+\alpha}{2}$, maximized by the water-filling policy $\pi^*(y)=[(f(y)-\nu)/(2\kappa_\alpha\lambda)]_+$. GRPO has $\kappa=\tfrac{K-2}{2(K-1)}$ and at $K=2$ converges to the single best answer for every $\lambda$.
  \item \emph{Distribution matching.} For $r_k=2t(y_k)-\tfrac{1}{K-1}\sum_{j\neq k}\ind[y_j=y_k]$ with target $t$, the group objective is $J=\|t\|^2-\|\pi-t\|^2$ and the $\alpha$-update converges to $\pi^*_\alpha=\Proj(2t/(1+\alpha))$, the Euclidean projection onto the simplex. \tg{} reaches $t$, REINFORCE reaches $\Proj(2t)$, and GRPO reaches $\Proj(2t\tfrac{K-1}{K-2})$, the mode of $t$ when $K=2$.
\end{enumerate}
\end{proposition}

GRPO therefore never reaches the target distribution, and the gap is not an optimization failure but the equilibrium of the game it plays. Retuning the reward coefficient is the natural remedy, and the identity says exactly when it works.

\begin{proposition}[When retuning suffices]\label{prop:retune}
Under \cref{cor:ladder}, GRPO with coefficient $\lambda'=\lambda\tfrac{m(K-1)}{K-m}$ has the expected update of the true gradient with coefficient $\lambda$, and no coefficient does when $K=m$. Under \cref{prop:pairwise}, if $\bar A$ is not an affine function of $\bar S$, no coefficient makes GRPO match the true gradient at any $K$.
\end{proposition}

Retuning thus helps only for symmetric kernels, only with the correct $K$-dependent multiplier ($3\times$ at $K=4$, $\tfrac73\times$ at $K=8$), and never at $K=2$, and quality-weighted diversity, the structure of DARLING, cannot be repaired at any $K$. With graded quality and embedding distances, the best GRPO coefficient on a grid recovers on average $11\%$, $89\%$ and $96\%$ of the optimal objective at $K=2$, $4$ and $8$ (\cref{app:screen}).

\paragraph{Beyond the plain update.}
These consequences are properties of the advantages themselves, so they hold for every loss built on them, including importance ratios, clipping, token-level averaging and several epochs per batch. At $K=2$ a symmetric pairwise term cancels in $r_1-r_2$, so GRPO computes exactly the advantages it would compute without the bonus, and standard-deviation normalization, which keeps only the sign of $r_1-r_2$, is equally blind. For a symmetric $m$-body term with $K>m$, the GRPO advantage with coefficient $\lambda$ is, group by group, a positive multiple of the \tg{} advantage with coefficient $\rho_m(K)\lambda$ (\cref{prop:samplewise}). The fixed points of \cref{prop:fixed} also survive clipping, since at a fixed point the ratio stays at one. In simulation, clipped updates with up to eight epochs per batch stay within $0.074$ of them in $\ell_1$, while the fixed points of GRPO and \tg{} lie $0.50$ to $1.10$ apart (\cref{app:ppo}).

\section{Team-GRPO}\label{sec:method}

\Cref{thm:identity} says that the gradient of the group objective counts each response's externality with weight $+1$. \tg{} does so with a baseline that remains independent of the response, so it keeps the variance reduction of group-relative training while following $\nabla_\theta J$. For a set $\mathcal S$ of responses let $R(\mathcal S)=\sum_{y_j\in\mathcal S}r(y_j;\mathcal S\setminus\{y_j\})$ be its total reward, computed as if $\mathcal S$ were the whole group. The \tg{} advantage is
\begin{equation}\label{eq:team}
  A_k^{\textsc{team}}=R(\gr)-\tfrac{K}{K-1}\,R(\gr\setminus\{y_k\}),
\end{equation}
which equals $K\big[\rbar(\gr)-\rbar(\gr\setminus\{y_k\})\big]$ with $\rbar(\mathcal S)=R(\mathcal S)/|\mathcal S|$. Each response is credited with $K$ times the change it causes in the group's mean reward. This is a difference reward in the sense of multi-agent credit assignment \citep{wolpert2001optimal,foerster2018counterfactual}, computed exactly and without a critic.

\begin{proposition}[Unbiasedness]\label{prop:unbiased}
For any exchangeable group reward and any $K\ge2$, $\E\big[\sum_k A^{\textsc{team}}_k\sth(y_k)\big]=K\nabla_\theta J$.
\end{proposition}
\begin{proof}
$R(\gr\setminus\{y_k\})$ does not depend on $y_k$ and contributes zero in expectation, and $\sum_k\E[R(\gr)\sth(y_k)]=\nabla_\theta\E[R(\gr)]=K\nabla_\theta J$.
\end{proof}

\paragraph{Cost.} For rewards built from pairwise distances, similarities or collisions, $R(\gr\setminus\{y_k\})$ follows from deleting row and column $k$ of the matrix the reward already uses (\cref{alg:team} in \cref{app:alg}). This is $O(K^2)$ arithmetic per group with no additional reward-model, embedding or judge calls. Rewards from a judge of the whole set need $K$ extra judge calls per group.

\paragraph{Two samples per prompt.} At $K=2$ the set $\gr\setminus\{y_k\}$ is a single response, for which a diversity term is undefined. Any baseline $b_k$ independent of $y_k$ keeps the estimator unbiased, so we use $A_k=r_1+r_2-b_k$ with $b_k$ an estimate of $\E[r_1+r_2\mid y_j]$ from the teammate $y_j$ and the rest of the batch. This lowers $\E[A^2]$ by $2.6\times$ relative to setting the missing diversity to zero (\cref{app:alg,app:novelty}).

\paragraph{Symmetric kernels in one line.} For a symmetric $m$-body interaction with $K>m$, \tg{} coincides sample by sample with RLOO whose interaction coefficient is multiplied by $m(K-1)/(K-m)$ (\cref{prop:samplewise}), which is $2(K-1)/(K-2)$ for collision penalties, frequency rewards and symmetric distance bonuses.

\paragraph{Relation to recent estimators.} Unbiased max@$k$ estimators \citep{walder2025pkpo,takashiro2026l2o} and SetPO's leave-one-out diversity bonus \citep{setpo2026} are \eqref{eq:team} for their rewards, and the uncentred witness estimator for distribution matching \citep{swys2026} is its symmetric-kernel form. The group-mean-centred witness is biased toward concentration by $\tfrac1K\nabla\|\pi\|^2$, as \cref{thm:identity} predicts (\cref{app:witness}). Poly-EPO \citep{orney2026polyepo} averages set advantages over subsets of size $n$ drawn from $N>n$ samples, so it estimates the same gradient but cannot run when the set is the whole group, including at $K=2$.

\paragraph{Normalization and variance.} As in Dr.\ GRPO and DARLING we do not divide by the group's standard deviation (\cref{app:std}). At matched scale the variance of \tg{} is $1.1$--$2.1$ times that of RLOO for $K\ge4$ (\cref{app:cpu}). The extra variance is the externality signal that GRPO cancels, and it shrinks with more prompts per step, whereas GRPO's bias, a relative error of $39$--$86\%$ in exact computations, does not.

\section{Experiments}\label{sec:exp}

We test whether the theory predicts where training converges, whether the tournament hurts across reward structures in current use, and whether \tg{} yields better models in distributional alignment and open-ended generation. Unless stated otherwise we fine-tune Qwen3.5-9B \citep{qwen35} with rank-16 LoRA \citep{hu2022lora} at temperature 1, with a sequence-level policy-gradient loss, a constant normalizer, no KL penalty and no standard-deviation normalization, identically for all estimators. Each batch is used for one on-policy update, where importance ratios equal one and the clipped surrogate of PPO and GRPO has exactly this gradient (normalization and clipping with several epochs are tested in \cref{app:std,app:ppo}). Confidence intervals are 95\%, from a paired bootstrap over prompts or a $t$-interval across seeds.

\subsection{The theory predicts where training converges}\label{sec:exp-theory}

\begin{figure*}[t]
  \centering
  \includegraphics[width=0.85\textwidth]{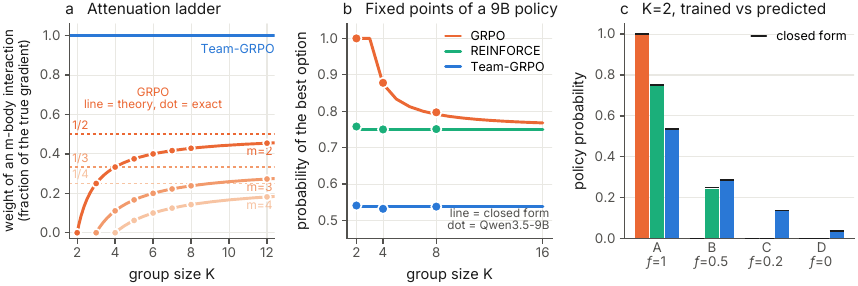}
  \vspace{-0.5em}
  \caption{\textbf{The theory is exact.}
  \textbf{(a)}~Weight GRPO gives an $m$-body symmetric interaction, relative to the true gradient. Lines are \cref{eq:retain}, dots are exact enumeration over random kernels, and dashed lines mark the limit $1/m$.
  \textbf{(b)}~Probability of the best option at the fixed point of the diversity bonus ($\lambda=1$). Lines are closed forms and dots are Qwen3.5-9B after 500 steps of LoRA training.
  \textbf{(c)}~Full policies at $K=2$, trained (bars) against closed form (black ticks). GRPO collapses onto the best option, REINFORCE reaches the Nash solution, and \tg{} reaches the optimum of the group reward.}
  \label{fig:theory}
  \vspace{-0.4em}
\end{figure*}

\paragraph{Exact enumeration.} Enumerating all answer tuples for small answer spaces gives every expectation exactly. For random exchangeable rewards with $K\in\{2,3,4,6,10\}$, \cref{thm:identity} holds to a relative error of $6\cdot10^{-15}$, and $g_1$ matches a finite-difference gradient of $J$. The attenuation ladder holds for $m\in\{2,3,4\}$ and $K\le12$ to $10^{-15}$ (\cref{fig:theory}a), and \cref{prop:pairwise} and the sample-wise identity for symmetric kernels hold to $6\cdot10^{-14}$ (\cref{app:cpu}).

\paragraph{A 9B policy with exact readout.} We let Qwen3.5-9B answer four-option prompts with a single letter, so that its policy can be read off exactly from the logits, and train it with the diversity bonus of \cref{prop:fixed}(a), with qualities $f=(1,0.5,0.2,0)$, on 128 prompts for 500 steps. All 12 runs with a closed-form prediction converge to the fixed point of their estimator's game, with $\ell_1$ error at most $0.018$ (\cref{fig:theory}b,c), and held-out prompts give the same policies (\cref{app:sandbox}). GRPO at $K=2$ puts all mass on the best option for $\lambda\in\{0,1,5\}$. \tg{} reaches the same diversified solution at $K=2$, $4$ and $8$, since its objective does not depend on $K$, and multiplying $\lambda$ by $2(K-1)/(K-2)=3$ moves GRPO at $K=4$ onto it, as \cref{prop:retune} prescribes. GRPO with standard-deviation normalization, which has no closed form, lands within $0.01$ per option of its numerical fixed point at $K=4$ and $8$ (\cref{app:std}).

\subsection{The tournament hurts across reward structures}\label{sec:exp-generality}

\Cref{tab:generality} computes the exact fixed point of each estimator for six group-reward structures from recent methods and reports the group objective there as a percentage of \tg{}'s. The ordering predicted by the $\alpha$-line, GRPO $\le$ REINFORCE $\le$ \tg{}, holds in all 18 settings, and GRPO is often far worse than no baseline at all. Its objective turns negative for distribution matching at $K=2$ and reaches $3\%$ of the optimum when only answers unique within the group are credited. The gap need not close with $K$. On the correctness-times-strategies objective of Poly-EPO, GRPO falls from $67\%$ at $K=2$ to $44\%$ at $K=8$ as it settles on a single correct strategy. GRPO matches \tg{} only where the optimum is an equilibrium of the tournament, as for pass@$K$, and standard-deviation normalization closes none of the gaps (\cref{app:std}).

\begin{table}[t]
  \centering
  \caption{\textbf{Group objective at each estimator's fixed point}, as a percentage of \tg{} (exact computation). Split credit divides a unit of reward among the members that earn it, so the group total is the set metric in parentheses. Shading marks the shortfall.}
  \label{tab:generality}
  \vspace{0.3em}
  \footnotesize
  \setlength{\tabcolsep}{2.9pt}
  \newcommand{\g}[2]{\cellcolor{cgrpo!#1}#2}
  \begin{tabular}{@{}lrrrrrr@{}}
    \toprule
     & \multicolumn{3}{c}{GRPO} & \multicolumn{3}{c}{REINFORCE} \\
    \cmidrule(lr){2-4}\cmidrule(lr){5-7}
    Group reward (used by) & $K{=}2$ & $4$ & $8$ & $2$ & $4$ & $8$ \\
    \midrule
    Distribution matching \scriptsize{(SWYS)}  & \g{60}{$-31$} & \g{24}{67} & \g{20}{75} & \g{15}{82} & \g{15}{82} & \g{15}{82} \\
    Quality $\times$ distance \scriptsize{(DARLING)} & \g{14}{83} & \g{9}{90} & \g{5}{96} & \g{3}{98} & \g{3}{98} & \g{3}{98} \\
    Correct $\times$ \#strategies \scriptsize{(Poly-EPO)} & \g{24}{67} & \g{34}{53} & \g{40}{44} & 100 & \g{3}{98} & \g{14}{83} \\
    Unique valid answers \scriptsize{(coverage)} & \g{55}{3} & \g{6}{94} & \g{42}{40} & 100 & \g{6}{94} & \g{42}{40} \\
    Split credit \scriptsize{(distinct@$K$)} & \g{28}{61} & 100 & 100 & 100 & 100 & 100 \\
    Split credit \scriptsize{(pass@$K$)} & 100 & 100 & 100 & 100 & 100 & 100 \\
    \bottomrule
  \end{tabular}
  \vspace{-0.6em}
\end{table}

\subsection{Distributional alignment}\label{sec:exp-opinion}

\paragraph{Setup.} GlobalOpinionQA \citep{durmus2024towards} pairs survey questions with the answer distributions of respondents in many countries. The model answers as a randomly selected adult of a given country with one option letter, and the target $t$ is that country's distribution of answers, a central goal of pluralistic alignment \citep{sorensen2024roadmap}. Splitting by question gives 32{,}372 training items and 1{,}154 test items from 510 unseen questions. The reward is that of \cref{prop:fixed}(b), whose group mean estimates $\|t\|^2-\|\pi-t\|^2$ without bias, so every estimator trains on exactly what we evaluate. Runs use 256 rollouts per step for 200 steps, and on test items we report an unbiased estimate of $\|\pi-t\|^2$ from 32 samples. Besides GRPO, GRPO (std) and REINFORCE, we compare with the centred witness estimator of \citet{swys2026} and with Poly-EPO \citep{orney2026polyepo} averaged over all pairs of the $K$ samples, whose optimum $t$ does not depend on the set size.  \tg{} uses the two-sample baseline of \cref{sec:method}, and for $K\ge4$ the symmetric-kernel form with a baseline from other questions.

\begin{table}[t]
  \centering
  \caption{\textbf{GlobalOpinionQA test error} $\|\pi-t\|^2$ after 200 steps (510 unseen questions, mean over seeds, base model $0.197$), with $K$ samples per question at 256 rollouts per step. Fixed points follow \cref{prop:fixed}(b) with $w_K=\tfrac{K-1}{K-2}$. Grey cells need no run, since at $K=2$ the centred witness gives exactly GRPO's advantages, REINFORCE has the same expected update at every $K$, and Poly-EPO needs more samples than its set size. Here DARLING's estimator is GRPO, GAPO's is GRPO (std), and SetPO's coincides with \tg{}. Bold is best per column.}
  \label{tab:opinion}
  \vspace{0.3em}
  \footnotesize
  \setlength{\tabcolsep}{3.4pt}
  \newcommand{\na}[1]{\textcolor{black!45}{\scriptsize #1}}
  \begin{tabular}{@{}llccc@{}}
    \toprule
    Estimator & Fixed point & $K{=}2$ & $K{=}4$ & $K{=}8$ \\
    \midrule
    GRPO & $\Proj(2w_Kt)$ & 0.491 & 0.181 & 0.179 \\
    REINFORCE & $\Proj(2t)$ & 0.131 & \multicolumn{2}{c}{\na{same update as $K{=}2$}} \\
    Centred witness & $\Proj(w_Kt)$ & \na{$=$ GRPO} & 0.096 & \textbf{0.073} \\
    Poly-EPO (pairs) & $t$ & \na{needs $K{>}2$} & 0.078 & 0.074 \\
    \textbf{\tg{}} & $t$ & \textbf{0.081} & \textbf{0.074} & 0.077 \\
    \bottomrule
  \end{tabular}
  \vspace{-0.8em}
\end{table}

\begin{figure*}[t]
  \centering
  \includegraphics[width=0.87\textwidth]{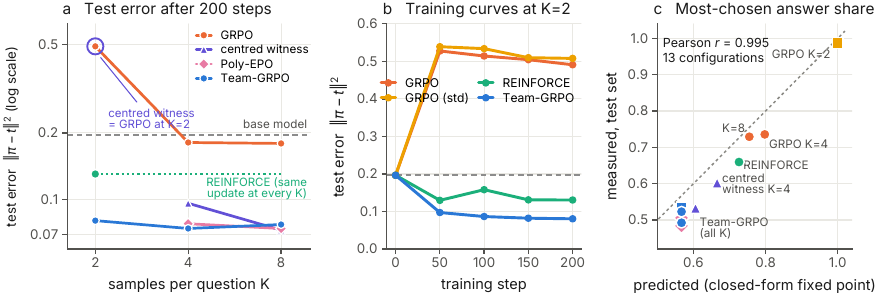}
  \vspace{-0.5em}
  \caption{\textbf{Distributional alignment on GlobalOpinionQA.}
  \textbf{(a)}~Squared error to the human distribution on unseen questions after 200 steps (log scale, means over seeds). Lines start or stop where the theory makes a run redundant (\cref{tab:opinion}). The bias of the centred witness shrinks with $K$ as predicted.
  \textbf{(b)}~Test error during training at $K=2$ (means over seeds). GRPO collapses within 50 steps, with or without standard-deviation normalization.
  \textbf{(c)}~Predicted against measured mass on the most-chosen answer for all 13 configurations with a closed-form fixed point, with Poly-EPO shown separately for 4 and 8 samples (dashed line is equality).}
  \label{fig:opinion}
  \vspace{-0.4em}
\end{figure*}

\paragraph{Results.} Every prediction holds (\cref{tab:opinion,fig:opinion}).
\emph{GRPO collapses at $K=2$.} At least 90\% of its samples agree on $97\%$ of test items, and its error of $0.49$ is two and a half times that of the untrained model. Standard-deviation normalization changes nothing ($0.508$, and $0.207$ at $K=4$).
\emph{\tg{} matches the target.} With the same samples it lowers the error to $0.08$, six times below GRPO (difference $-0.40$ $[-0.43,-0.38]$), and REINFORCE lands in between ($0.13$), where the $\alpha$-line places it.
\emph{Larger groups do not rescue GRPO.} At $K=4$ and $8$ its error stays at $0.18$, $2.3$--$2.4$ times that of \tg{}, because its fixed point never reaches $t$.
\emph{Where the theory predicts a tie, there is one.} Poly-EPO estimates the same gradient as \tg{} and ends at similar errors wherever it can run ($0.078$ against $0.074$ at $K=4$, $0.074$ against $0.077$ at $K=8$). The centred witness is biased by $\tfrac1K\nabla\|\pi\|^2$, which predicts a clear gap at $K=4$ and almost none at $K=8$. It ends at $0.096$ against $0.074$ at $K=4$ (difference $-0.022$ $[-0.029,-0.015]$), and at $K=8$, where its fixed point lies within $0.0033$ of $t$ in squared error, the three estimators finish within $0.005$ of each other, no more than the gap between different seeds of \tg{}.
\emph{The predictions are quantitative.} Across all 13 configurations with a closed-form fixed point, predicted and measured concentration agree with Pearson $r=0.995$ (\cref{fig:opinion}c), with measured values on average $0.05$ below the predictions. \tg{} also wins at equal distance from the base model (\cref{app:opinion}).

\subsection{Open-ended generation}\label{sec:exp-novelty}

\begin{figure*}[t]
  \centering
  \includegraphics[width=0.89\textwidth]{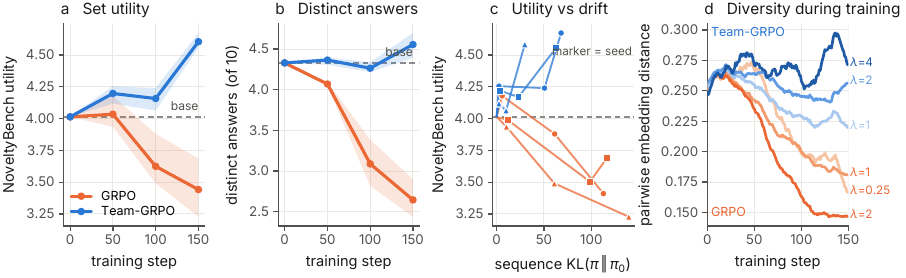}
  \vspace{-0.5em}
  \caption{\textbf{Open-ended generation on NoveltyBench} ($K=2$, $\lambda=2$ unless noted).
  \textbf{(a,b)}~Set utility and number of distinct answers among 10 samples. Lines are means over three seeds and shading spans the seeds.
  \textbf{(c)}~Utility against sequence-level KL divergence from the base model, per seed.
  \textbf{(d)}~Mean pairwise embedding distance within training groups (smoothed). GRPO collapses faster as the diversity weight $\lambda$ grows, while \tg{} becomes more diverse.}
  \label{fig:novelty}
  \vspace{-0.4em}
\end{figure*}

\paragraph{Setup.} NoveltyBench \citep{zhang2025noveltybench} scores 10 samples per open-ended prompt. \emph{Distinct} counts the equivalence classes among them and \emph{utility} sums the quality of novel samples with a patience discount, so both are set-level objectives of the kind $J$ describes. We train on 4{,}000 WildChat \citep{zhao2024wildchat} prompts chosen with the benchmark's recipe and disjoint from its test set, and evaluate on 400 prompts with the official classifier and scorer, whose published numbers our pipeline reproduces. The reward is $r_k=q_k(1+\lambda D_k)$, with $q_k$ the calibrated score of Skywork-Reward-V2-Llama-3.1-8B \citep{liu2025skyworkv2} and $D_k$ the mean E5 embedding distance \citep{wang2022e5} to the groupmates, the DARLING structure that no retuning can fix (\cref{prop:retune}). We use $\lambda=2$, $K\in\{2,8\}$, 64 rollouts per step, 150 steps at learning rate $5\cdot10^{-5}$, and three seeds for each group size.

\begin{table}[t]
  \centering
  \caption{\textbf{NoveltyBench performance} (benchmark scorer, 400 prompts, three seeds). $\Delta$ is \tg{} minus GRPO with 95\% CIs, and at $K=8$ matched KL compares checkpoints within the common observed KL range. Bold is the best trained model on set metrics, and 1st sample scores one response. LLM-judge results are in \cref{tab:judge}.}
  \label{tab:novelty}
  \vspace{0.3em}
  \footnotesize
  \setlength{\tabcolsep}{2.6pt}
  \resizebox{\columnwidth}{!}{%
  \begin{tabular}{@{}lccc@{}}
    \toprule
    & Utility $\uparrow$ & Distinct $\uparrow$ & 1st sample \\
    \midrule
    \multicolumn{4}{@{}l}{\emph{$K=2$}} \\
    Base model & 4.01 & 4.33 & 7.58 \\
    GRPO ($\lambda{=}2$) & 3.44 & 2.65 & 9.04 \\
    GRPO (best $\lambda{=}1$) & 3.83 & 3.19 & 9.07 \\
    \textbf{\tg{}} ($\lambda{=}2$) & \textbf{4.61} & \textbf{4.56} & 8.51 \\
    $\Delta$, same step & $+1.16$ {\tiny$[0.52,1.80]$} & $+1.91$ {\tiny$[1.01,2.81]$} & \\
    $\Delta$, matched KL & $+0.84$ {\tiny$[0.79,0.90]$} & $+1.12$ {\tiny$[0.46,1.79]$} & \\
    \midrule
    \multicolumn{4}{@{}l}{\emph{$K=8$ (same base model)}} \\
    GRPO ($\lambda{=}2$) & 4.12 & 3.51 & 9.07 \\
    \textbf{\tg{}} ($\lambda{=}2$) & \textbf{4.68} & \textbf{4.55} & 8.70 \\
    $\Delta$, same step & $+0.56$ {\tiny$[0.32,0.81]$} & $+1.04$ {\tiny$[0.99,1.09]$} & \\
    $\Delta$, matched KL & $+0.62$ {\tiny$[0.30,0.95]$} & $+0.72$ {\tiny$[0.34,1.11]$} & \\
    \bottomrule
  \end{tabular}}
  \vspace{-0.8em}
\end{table}

\paragraph{GRPO trains a quality-only policy.} At $K=2$ the GRPO advantage \eqref{eq:k2adv} ignores diversity, and in all three seeds the trained model has fewer distinct answers and lower utility than the base model. Averaged over seeds, distinct answers fall from $4.33$ to $2.65$ and utility from $4.01$ to $3.44$, while first-sample quality rises from $7.58$ to $9.04$ (\cref{tab:novelty}, \cref{fig:novelty}a,b). This rise is the predicted signature of the failure, not a strength. A policy that sees only quality moves its mass onto its single best answer, which raises the score of any one sample while the set collapses. \tg{} makes the trade the reward asks for, giving up $0.53$ points of single-sample quality to win on both set metrics.

\paragraph{\tg{} optimizes the set.} Under the same reward, \tg{} keeps the model diverse ($4.56$ distinct answers) and raises utility to $4.61$, significantly above both the base model and GRPO ($+1.16$ $[0.52,1.80]$ over GRPO). GRPO drifts further from the base model (sequence KL $123$ against $54$), so we also evaluate GRPO at the KL each \tg{} run reaches at step 150, where \tg{} still leads by $+0.84$ $[0.79,0.90]$ in utility and $+1.12$ $[0.46,1.79]$ in distinct answers (\cref{fig:novelty}c). Single prompts show the difference plainly. Asked to pick a random prime between 1 and 50, GRPO answers 23 in seven of ten samples, while \tg{} names six different primes (\cref{tab:examples}).

\paragraph{More diversity weight makes GRPO worse.} \Cref{eq:k2adv} predicts that a larger $\lambda$ cannot help GRPO at $K=2$ and accelerates its collapse. Over the last 25 training steps, GRPO's mean pairwise distance drops from $0.193$ to $0.186$ to $0.147$ as $\lambda$ goes from $0.25$ to $1$ to $2$, whereas that of \tg{} rises from $0.231$ to $0.249$ to $0.293$ for $\lambda=1,2,4$ (\cref{fig:novelty}d). The best GRPO coefficient ($\lambda=1$) reaches only $3.83$ utility, below the untrained model.

\paragraph{Larger groups preserve the advantage.} With eight samples per prompt at the same 64-rollout budget, GRPO improves to $4.12$ utility and $3.51$ distinct answers, while \tg{} reaches $4.68$ and $4.55$ (\cref{tab:novelty}). The paired differences across three seeds are $+0.56$ $[0.32,0.81]$ in utility and $+1.04$ $[0.99,1.09]$ in distinct answers, and $+0.62$ $[0.30,0.95]$ and $+0.72$ $[0.34,1.11]$ at matched KL. Larger groups narrow the gap, as the attenuation ladder predicts, while \tg{} retains an advantage on both set metrics.

\paragraph{An independent judge agrees.} The training reward model and the benchmark scorer both come from the Skywork family, so we also apply the NoveltyBench v1.1 protocol, in which an external language model (Claude Sonnet 5.5) groups and scores the samples, to a random subset of 100 prompts (\cref{tab:judge}). \tg{} beats GRPO in utility and distinct answers at both group sizes, by $+0.61$ and $+2.09$ at $K=2$ and by $+0.26$ and $+1.34$ at $K=8$, all with intervals excluding zero, and it is the only trained model above the base model on both set metrics. The differences remain positive at matched KL, and as under the benchmark scorer GRPO's first sample scores higher while its set scores lower.


\section{Related Work}\label{sec:related}

\paragraph{Baselines and credit assignment.} Baselines reduce variance without bias as long as they do not depend on the action \citep{williams1992simple,greensmith2004variance}. Leave-one-out and group-mean baselines \citep{kool2019buy,ahmadian2024back,shao2024deepseekmath} made critic-free training standard for language models, and later work examined their normalization and length bias \citep{liu2025understanding,yu2025dapo,vojnovic2025alignment}. Action-dependent baselines need a correction to stay unbiased \citep{tucker2018mirage,wu2018variance}, and the bias we characterize is of this kind, entering through the groupmates' rewards. Counterfactual baselines \citep{foerster2018counterfactual} and difference rewards \citep{wolpert2001optimal} credit each agent with its marginal contribution, and \tg{} applies this idea to samples of one policy, where the counterfactual is exact. Multi-agent GRPO variants with baselines from independent episodes \citep{liu2025magrpo} are unaffected. Selection in finite populations maximizes relative payoff and can reward spite \citep{schaffer1988evolutionarily,nowak2004emergence,hamilton1970selfish,vega1997evolution}, as tournament contracts do \citep{lazear1981rank,lazear1989pay}.

\paragraph{Group-dependent rewards.} Diversity methods reward semantic distinctness, embedding distance or rarity within the group \citep{li2025jointly,song2025outcome,dsdr2026,anschel2025gapo,shi2026spader}. Distributional and pluralistic alignment ask a policy to match a population of views \citep{durmus2024towards,sorensen2024roadmap,swys2026}, and set-level objectives target pass@$k$ \citep{walder2025pkpo,takashiro2026l2o}, strategy coverage \citep{orney2026polyepo,hamid2026spiral}, marginal diversity \citep{setpo2026} or within-group preferences \citep{wang2025prefgrpo}. Most feed the group-dependent reward into the standard advantage, and GAPO argues explicitly that this is unbiased because the reward does not contain $\theta$ \citep{anschel2025gapo}. Our analysis shows that the estimator then plays a tournament, and it gives one correction that also holds in the small groups favored for efficiency \citep{twogrpo2025}.

\section{Conclusion}\label{sec:conclusion}

Group-relative baselines were introduced to reduce variance. When the reward depends on the group they also change the game, from a team effort into a tournament in which every response is rewarded for beating its groupmates. The change hides diversity from two-sample training, keeps less than half of any pairwise bonus, drives distribution matching toward the majority answer, and leaves a bias that larger groups do not remove. Counting each response's externality with the right sign restores the true gradient at every group size, without additional model calls for pairwise rewards.

\section*{Impact Statement}
This work concerns how reinforcement-learning objectives for language models are optimized when rewards depend on several sampled responses at once. Rewards of this kind are used to keep models diverse, to represent the spread of human opinion rather than its majority, and to cover many valid answers. We show that the standard estimator quietly works against these goals and provide a corrected estimator. A model trained toward a population's opinion distribution reproduces that distribution, including its disagreements, so practitioners should choose target populations with care. We see no other societal consequence that needs to be highlighted here.

\bibliography{refs}
\bibliographystyle{icml2026}

\newpage
\appendix
\onecolumn
\section*{Appendix}

\Cref{app:proofs} proves the results of \cref{sec:theory,sec:method}, and \cref{app:alg} gives the algorithm. \Cref{app:cpu,app:screen,app:std,app:ppo} report the exact computations behind \cref{sec:exp-theory,sec:exp-generality}, the limits of retuning, the effect of standard-deviation normalization, and the effect of clipping with several epochs per batch. \Cref{app:methods} applies the identity to published methods. \Cref{app:sandbox,app:opinion,app:novelty} give the details and additional results of the three model-training experiments.

\section{Proofs}\label{app:proofs}

Throughout, $y_1,\dots,y_K$ are i.i.d.\ draws from $\pth$, $r_k=r(y_k;\{y_j\}_{j\neq k})$ is exchangeable, and $\sth(y)=\nabla_\theta\log\pth(y)$ satisfies $\E[\sth(y)]=0$. Because the samples are i.i.d.\ and the reward is exchangeable, the joint law of $\big(y_k,r_k,\{r_j\}_{j\neq k}\big)$ does not depend on $k$, so $\own$ and $\ext$ in \eqref{eq:ownext} may be defined with any index in place of $1$. We write $h_\alpha=\own+\alpha\,\ext$, so that $g_\alpha=\E[h_\alpha(y)\,\sth(y)]$.

\subsection{Proof of \cref{thm:identity}}
\emph{True gradient.} The reward does not depend on $\theta$ directly, so
\begin{equation*}
  \nabla_\theta J=\tfrac1K\nabla_\theta\E\Big[\sum_j r_j\Big]=\tfrac1K\sum_{k=1}^K\E\Big[\Big(\sum_j r_j\Big)\sth(y_k)\Big].
\end{equation*}
For each $k$, $\E[r_k\,\sth(y_k)]=\E\big[\E[r_k\mid y_k]\,\sth(y_k)\big]=\E[\own(y_k)\sth(y_k)]$ and $\sum_{j\neq k}\E[r_j\,\sth(y_k)]=\E\big[\sum_{j\neq k}\E[r_j\mid y_k]\,\sth(y_k)\big]=\E[\ext(y_k)\sth(y_k)]$. Each of the $K$ terms equals $g_1$, hence $\nabla_\theta J=g_1$.

\emph{REINFORCE.} With $A_k=r_k-b_k$ and $b_k$ independent of $y_k$ (a constant, a function of other prompts, or of $y_{-k}$), $\E[b_k\,\sth(y_k)]=\E\big[b_k\,\E[\sth(y_k)\mid b_k]\big]=0$. Each response contributes $\E[\own\,\sth]=g_0$, and the sum is $Kg_0$.

\emph{RLOO and GRPO.} With $A_k=r_k-\tfrac{1}{K-1}\sum_{j\neq k}r_j$, the same conditioning gives $\E[A_k\,\sth(y_k)]=\E[\own\,\sth]-\tfrac{1}{K-1}\E[\ext\,\sth]=g_{-1/(K-1)}$, and summing over $k$ gives $Kg_{-1/(K-1)}$. GRPO's advantage is $r_k-\rbar=\tfrac{K-1}{K}\big(r_k-\tfrac{1}{K-1}\sum_{j\neq k}r_j\big)$, which multiplies the result by $(K-1)/K$. \hfill$\square$

\subsection{The game behind each estimator}

\begin{proposition}\label{prop:game}
Give each of $K$ players its own parameters $\theta_k$ and the payoff $U^\alpha_k=r_k+\alpha\sum_{j\neq k}r_j$. At the symmetric profile $\theta_1=\dots=\theta_K=\theta$, the gradient of player $k$'s expected payoff with respect to $\theta_k$ is $g_\alpha(\theta)$. For a softmax policy over a finite answer set with logits $\theta$, the expected $\alpha$-update on the logits is $K\,\pi(y)\big(h_\alpha(y)-\E_\pi h_\alpha\big)$, which vanishes if and only if $h_\alpha$ is constant on the support of $\pi$, the first-order condition for $\pi$ to be a symmetric equilibrium of the game with payoffs $U^\alpha$. For $\alpha=-1/(K-1)$ the payoffs satisfy $\sum_k U_k=0$.
\end{proposition}
\begin{proof}
$\nabla_{\theta_k}\E[U^\alpha_k]=\E[U^\alpha_k\,s_{\theta_k}(y_k)]$, and the computation in the proof of \cref{thm:identity} gives $\E[(r_k+\alpha\sum_{j\neq k}r_j)\sth(y_k)]=g_\alpha$ at the symmetric profile. For softmax logits, $\sth(y)=e_y-\pi$, so $g_\alpha=\sum_y\pi(y)h_\alpha(y)(e_y-\pi)$, whose $y$-th coordinate is $\pi(y)(h_\alpha(y)-\E_\pi h_\alpha)$. Finally $\sum_k\big(r_k-\tfrac{1}{K-1}\sum_{j\neq k}r_j\big)=\sum_k r_k-\tfrac{K-1}{K-1}\sum_j r_j=0$.
\end{proof}
The logit dynamics $\dot\theta\propto\pi\odot(h_\alpha-\E_\pi h_\alpha)$ are a replicator-type dynamics with fitness $h_\alpha$ \citep{taylor1978evolutionary}. For $\alpha=-1/(K-1)$ the fitness of a strategy is its payoff relative to the mean payoff of the other $K-1$ members, the quantity that drives selection in a population of size $K$ \citep{schaffer1988evolutionarily}.

\subsection{Proof of \cref{cor:ladder}}
Let $c_k=\binom{K-1}{m-1}^{-1}\sum_{S}\phi(y_k,y_S)$, the sum running over subsets $S$ of $m-1$ groupmates of $k$. Then $\own_c(y)=\E[c_1\mid y_1=y]=\Phi(y)$. For $j\ge2$, a fraction $\tfrac{m-1}{K-1}$ of the subsets defining $c_j$ contain index $1$. For each of them, symmetry of $\phi$ gives $\E[\phi(y_j,y_S)\mid y_1=y]=\Phi(y)$, while the remaining subsets contribute a constant. Hence $\E[c_j\mid y_1=y]=\tfrac{m-1}{K-1}\Phi(y)+\text{const}$ and, summing over the $K-1$ groupmates, $\ext_c(y)=(m-1)\Phi(y)+\text{const}$. Constants vanish against the score, which proves \eqref{eq:ladder}. Setting $\alpha=1$ and $\alpha=-1/(K-1)$ gives the weights $m$ and $(K-m)/(K-1)$, and their ratio is \eqref{eq:retain}, which is below $1/m$ because $K-m<K-1$.

If $\sum_k c_k\equiv C$, then $\ext_c(y)=\E[C-c_1\mid y_1=y]=C-\own_c(y)$, so the interaction weight is $1-\alpha$, equal to $K/(K-1)$ for GRPO and $0$ for the true gradient. If $r_k=R(y_1,\dots,y_K)$ for every $k$, then $\ext=(K-1)\own$ and the weight is $1+\alpha(K-1)$, which is $0$ for GRPO. \hfill$\square$

\subsection{Proof of \cref{prop:pairwise}}
$\own_M(y)=\tfrac{1}{K-1}\sum_{j\neq1}\E[M(y,y_j)]=\bar S(y)+\bar A(y)$. For $j\ge2$, the only term of $r_j$ that involves $y_1$ is $\tfrac{1}{K-1}M(y_j,y_1)$, whose conditional mean is $\tfrac{1}{K-1}\E_{y'}M(y',y)=\tfrac{1}{K-1}\big(\bar S(y)-\bar A(y)\big)$. Summing over $j\ge2$ gives $\ext_M=\bar S-\bar A+\text{const}$, and $\own_M+\alpha\,\ext_M=(1+\alpha)\bar S+(1-\alpha)\bar A$ up to a constant.

\emph{The rotational component.} On a finite answer space, $\bar S=S\pi$ and $\bar A=A\pi$ with $S$ symmetric and $A$ antisymmetric matrices. The interaction field $\pi\mapsto(1+\alpha)S\pi+(1-\alpha)A\pi$ has Jacobian $(1+\alpha)S+(1-\alpha)A$. A gradient field has a symmetric Jacobian, so the field is a gradient (of $\tfrac{1+\alpha}{2}\pi^\top S\pi$) only when $\alpha=1$ or when $A$ vanishes on the tangent space of the simplex. For GRPO with $K>2$ the antisymmetric part enters with weight $K/(K-1)$, and the update is not the gradient of any function of $\pi$. \hfill$\square$

\subsection{Proof of \cref{prop:fixed}}
For a softmax policy let $D_\pi=\mathrm{diag}(\pi)-\pi\pi^\top$, which is symmetric positive semidefinite. By \cref{prop:game} the expected update is $\dot\theta=K D_\pi h_\alpha$, so $\dot\pi=D_\pi\dot\theta=K D_\pi^2h_\alpha$. If $h_\alpha=\nabla_\pi\Psi$ up to an additive constant, then $\dot\Psi=h_\alpha^\top\dot\pi=K\|D_\pi h_\alpha\|^2\ge0$, so $\Psi$ increases along the flow and its stationary points are those where $h_\alpha$ is constant on the support.

(a) Here $\phi(y,y')=\ind[y\neq y']$ is symmetric with $\Phi(y)=1-\pi(y)$, so by \cref{cor:ladder} $h_\alpha(y)=f(y)+\lambda(1+\alpha)(1-\pi(y))$, which is the gradient on the simplex of $\Psi_\alpha=\E_\pi f+\kappa_\alpha\lambda(1-\|\pi\|^2)$ with $\kappa_\alpha=\tfrac{1+\alpha}{2}$, up to a constant. For $\kappa_\alpha\lambda>0$, $\Psi_\alpha$ is strictly concave and its KKT conditions $f(y)-2\kappa_\alpha\lambda\pi(y)=\nu$ on the support and $f(y)\le\nu$ off it give the water-filling solution. For GRPO, $\kappa=\tfrac12\big(1-\tfrac{1}{K-1}\big)=\tfrac{K-2}{2(K-1)}$, which is zero at $K=2$, where $\Psi$ is linear and maximized by the best answer.

(b) The collision kernel $\phi=-\ind[y=y']$ gives $h_\alpha(y)=2t(y)-(1+\alpha)\pi(y)$ up to a constant, the gradient of $\Psi_\alpha=2\langle t,\pi\rangle-\tfrac{1+\alpha}{2}\|\pi\|^2=-\tfrac{1+\alpha}{2}\big\|\pi-\tfrac{2t}{1+\alpha}\big\|^2+\text{const}$, whose maximizer on the simplex is $\Proj(2t/(1+\alpha))$. The group objective is $\E[r_k]=2\langle t,\pi\rangle-\|\pi\|^2=\|t\|^2-\|\pi-t\|^2$. For GRPO, $1+\alpha=(K-2)/(K-1)$. At $K=2$ the potential is $2\langle t,\pi\rangle$ and is maximized at the mode of $t$. \hfill$\square$

\subsection{Proof of \cref{prop:retune}}
Under \cref{cor:ladder}, GRPO with coefficient $\lambda'$ has $h=f+\lambda'\tfrac{K-m}{K-1}\Phi$ and the true gradient has $h=f+\lambda m\Phi$. They coincide if and only if $\lambda'(K-m)/(K-1)=\lambda m$, which has a solution if and only if $K>m$. Under \cref{prop:pairwise}, GRPO with $\lambda'$ has $h=f+\lambda'\big[\tfrac{K-2}{K-1}\bar S+\tfrac{K}{K-1}\bar A\big]$ and the true gradient has $h=f+2\lambda\bar S$. Equality up to a constant requires $\lambda'\tfrac{K}{K-1}\bar A$ to be an affine function of $\bar S$, which fails whenever $\bar A$ is not, unless $\lambda'=0$. A positive rescaling of the whole update does not help, because it would also rescale $f$. \hfill$\square$

\subsection{An unbiased estimator for every point of the line}\label{app:family}
Let $r_j^{(-k)}=r(y_j;\gr\setminus\{y_j,y_k\})$ be the reward of $y_j$ recomputed without $y_k$, which does not depend on $y_k$. For any $\alpha$,
\begin{equation*}
  A_k^{\alpha}=r_k-\tfrac{1}{K-1}\sum_{j\neq k}r_j^{(-k)}+\alpha\sum_{j\neq k}\big(r_j-r_j^{(-k)}\big)
\end{equation*}
satisfies $\E[\sum_k A_k^\alpha\sth(y_k)]=Kg_\alpha$, because the terms in $r_j^{(-k)}$ are independent of $y_k$ and the remaining terms contribute $\own+\alpha\,\ext$ as in the proof of \cref{thm:identity}. For $\alpha=1$ the advantage telescopes to $\sum_j r_j-\tfrac{K}{K-1}\sum_{j\neq k}r_j^{(-k)}$, which is \eqref{eq:team}. The family separates the two ingredients of the correction. Recomputing the comparators without $y_k$ removes the hidden dependence of the baseline on $y_k$, and the externality $r_j-r_j^{(-k)}$ is then added back with the chosen weight. $\alpha=0$ gives an unbiased estimator of the Nash direction, for a designer who wants each response credited with its groupmates held fixed. Either intention thus has an exact estimator, and GRPO, with its negative weight, serves neither.

\subsection{Sample-wise identity for symmetric kernels}\label{app:samplewise}
\begin{proposition}\label{prop:samplewise}
Under \cref{cor:ladder} with $K>m$, the \tg{} advantage \eqref{eq:team} equals, for every group, the RLOO advantage computed with coefficient $\lambda'=\lambda\,m(K-1)/(K-m)$.
\end{proposition}
\begin{proof}
Write $\phi_T=\phi(y_T)$ for each $m$-subset $T$ of the group, $N_1=\binom{K-1}{m-1}$ and $N_2=\binom{K-2}{m-1}$, so $N_2/N_1=(K-m)/(K-1)$. Both advantages are linear in $(f_j)_j$ and $(\phi_T)_T$. The $f$-parts agree, since both equal $f_k-\tfrac{1}{K-1}\sum_{j\neq k}f_j$. In the RLOO advantage the coefficient of $\phi_T$ is $\lambda'\tfrac{K-m}{(K-1)N_1}$ if $k\in T$ and $-\lambda'\tfrac{m}{(K-1)N_1}$ otherwise. In \eqref{eq:team}, $R(\gr)$ gives every $\phi_T$ the coefficient $\lambda m/N_1$, and $\tfrac{K}{K-1}R(\gr\setminus\{y_k\})$ gives $\lambda\tfrac{K}{K-1}\tfrac{m}{N_2}=\lambda\tfrac{Km}{(K-m)N_1}$ to the subsets not containing $k$. The coefficients are therefore $\lambda m/N_1$ for $k\in T$ and $-\lambda m^2/((K-m)N_1)$ otherwise, which match the RLOO coefficients exactly when $\lambda'=\lambda m(K-1)/(K-m)$.
\end{proof}

\subsection{Bias of the centred witness estimator}\label{app:witness}
The witness estimator for distribution matching uses $w_k=2t(y_k)-\tfrac{2}{K-1}\sum_{j\neq k}\ind[y_j=y_k]$ \citep{swys2026}. Without centring, $w_k$ doubles the collision term of $r_k$, and by \cref{prop:samplewise} (in expectation) it follows the true gradient. With the group mean subtracted, the term $-\tfrac1K\sum_{j\neq k}w_j$ adds $-\tfrac1K\ext_w$, and $\ext_w(y)=-2\pi(y)+\text{const}$. The expected update is therefore proportional to
\begin{equation*}
  h(y)=\tfrac{K-1}{K}\big(2t(y)-2\pi(y)\big)+\tfrac{2}{K}\pi(y),
\end{equation*}
the true gradient plus $\tfrac1K\nabla_\pi\|\pi\|^2$, a push toward concentration. Its fixed point is $\Proj\big(t\,\tfrac{K-1}{K-2}\big)$, which is $\Proj(\tfrac32t)$ at $K=4$ and $\Proj(\tfrac76t)$ at $K=8$ (\cref{tab:opinion}).

\section{Implementing \tg{}}\label{app:alg}

\Cref{alg:team} computes the \tg{} advantages of one group. For rewards built from a pairwise matrix, $R(\gr\setminus\{y_k\})$ reuses the matrix with row and column $k$ deleted, so no further reward-model, embedding or judge calls are needed. At $K=2$ the leave-one-out set holds a single response, and we use the baseline $b_k$ of \cref{sec:method}, an estimate of $\E[r_1+r_2\mid y_j]$ for the teammate $y_j$. We evaluate the reward at the batch-mean distance, with the teammate's quality and, for the missing response, a quality shrunk from the batch mean toward the teammate's by the within-prompt quality correlation of previous steps (\cref{tab:k2var}).

\begin{algorithm}[h]
  \caption{\tg{} advantages for one prompt}
  \label{alg:team}
  \begin{algorithmic}[1]
    \STATE Sample $y_1,\dots,y_K\sim\pth$ and compute the per-response scores and pairwise quantities the reward uses.
    \STATE $R(\gr)\leftarrow\sum_j r_j$
    \FOR{$k=1$ \textbf{to} $K$}
      \STATE $R(\gr\setminus\{y_k\})\leftarrow$ rewards recomputed with row and column $k$ of the pairwise matrix removed
      \STATE $A_k\leftarrow R(\gr)-\tfrac{K}{K-1}R(\gr\setminus\{y_k\})$ \hfill ($K\ge3$)
      \STATE $A_k\leftarrow R(\gr)-b_k$, $b_k$ from the teammate and the batch \hfill ($K=2$)
    \ENDFOR
    \STATE Use $A_k$ in place of the GRPO advantage.
  \end{algorithmic}
\end{algorithm}

\section{Exact computations}\label{app:cpu}

We represent a policy over $n\le6$ answers by softmax logits and compute every expectation by enumerating all $n^K$ answer tuples (or one representative per multiset, weighted by its multinomial coefficient). Advantages, expected updates and second moments are therefore exact up to floating-point error. Fixed points are approximated by deterministic ascent on the exact expected update.

\paragraph{Identities.} \Cref{tab:identity} reports the largest relative error of each identity over randomly generated exchangeable rewards, defined by random tables over (own answer, multiset of groupmates' answers). The ladder coefficients of \cref{cor:ladder} were checked for symmetric, all-distinct and constant-sum kernels with $m\in\{2,3,4\}$ and $K\in\{m,\dots,8,12\}$, with largest error $9\cdot10^{-16}$. \Cref{prop:pairwise} was checked for random $S$, $A$ and $\pi$ at $K\in\{2,3,4,8\}$ and $\alpha\in\{-\tfrac{1}{K-1},0,1\}$ (largest error $6\cdot10^{-14}$), and \cref{prop:samplewise} for $m\in\{2,3,4\}$, $K\le8$ and four values of $\alpha$ on every answer tuple (largest error $1.1\cdot10^{-15}$).

\begin{table}[h]
  \centering
  \caption{Largest relative error of \cref{thm:identity} in exact enumeration over random exchangeable rewards.}
  \label{tab:identity}
  \small
  \begin{tabular}{@{}lccccc@{}}
    \toprule
    Identity & $K=2$ & $3$ & $4$ & $6$ & $10$ \\
    \midrule
    $\E\hat g_{\textsc{reinforce}}=Kg_0$ & $3.1\cdot10^{-16}$ & $3.7\cdot10^{-16}$ & $1.0\cdot10^{-15}$ & $4.4\cdot10^{-16}$ & $2.6\cdot10^{-15}$ \\
    $\E\hat g_{\textsc{rloo}}=Kg_{-1/(K-1)}$ & $3.2\cdot10^{-16}$ & $2.2\cdot10^{-16}$ & $2.6\cdot10^{-16}$ & $4.9\cdot10^{-16}$ & $2.1\cdot10^{-15}$ \\
    $\E\hat g_{\textsc{grpo}}=\tfrac{K-1}{K}\E\hat g_{\textsc{rloo}}$ & $1.4\cdot10^{-16}$ & $1.9\cdot10^{-16}$ & $2.7\cdot10^{-16}$ & $3.3\cdot10^{-16}$ & $7.0\cdot10^{-16}$ \\
    $g_1=\nabla J$ (finite differences) & $6.1\cdot10^{-11}$ & $2.0\cdot10^{-10}$ & $1.0\cdot10^{-10}$ & $1.1\cdot10^{-10}$ & $9.3\cdot10^{-11}$ \\
    \bottomrule
  \end{tabular}
\end{table}

\paragraph{Variance.} \Cref{tab:variance} compares the three estimators at a random policy for the diversity bonus of \cref{prop:fixed}(a) with $\lambda=1$. Variances are normalized by the squared projection of each expected update on a common direction, so that differences in step size do not count. \tg{} is the only unbiased estimator of $\nabla J$ in the table. Its variance is within a factor of about two of RLOO for $K\ge4$, while the bias of GRPO is a large fraction of the gradient at every $K$. The comparison favours RLOO by construction, because RLOO's lower variance comes from cancelling the externality term, which is part of $\nabla J$. Averaging over $B$ prompts divides the variance of \tg{} by $B$ and leaves the bias of GRPO untouched, so the error of \tg{} vanishes with more prompts per step while that of GRPO does not. At $K=3$ the variance is larger ($5.3$) because the counterfactual group $\gr\setminus\{y_k\}$ has only two members, and it falls quickly as $K$ grows.

\begin{table}[h]
  \centering
  \caption{Relative error of the expected update with respect to $\nabla J$, and variance at matched scale relative to RLOO, at a random policy (exact).}
  \label{tab:variance}
  \small
  \begin{tabular}{@{}lcccccc@{}}
    \toprule
    & \multicolumn{2}{c}{\tg{}} & \multicolumn{2}{c}{REINFORCE} & \multicolumn{2}{c}{GRPO / RLOO} \\
    \cmidrule(lr){2-3}\cmidrule(lr){4-5}\cmidrule(lr){6-7}
    $K$ & error & variance & error & variance & error & variance \\
    \midrule
    3 & $1\cdot10^{-10}$ & 5.33 & 0.49 & 6.45 & 0.78 & 1 \\
    4 & $1\cdot10^{-10}$ & 1.69 & 0.43 & 2.26 & 0.64 & 1 \\
    8 & $4\cdot10^{-10}$ & 2.11 & 0.31 & 4.44 & 0.39 & 1 \\
    16 & $1\cdot10^{-9}$ & 1.07 & 0.80 & 6.55 & 0.86 & 1 \\
    \bottomrule
  \end{tabular}
\end{table}

\section{Retuning the coefficient}\label{app:screen}

\Cref{prop:retune} says when a rescaled coefficient can stand in for \tg{}. \Cref{tab:screen} tests this on three reward families by exact computation. For each, we give GRPO the default coefficient and the best coefficient on a grid, and report the group objective at its fixed point as a percentage of \tg{}'s. For the binary same-class kernel of DARLING, retuning closes the gap at $K\ge4$, but only with the unusual, $K$-dependent multipliers that \cref{prop:retune} predicts ($\kappa=3$ at $K=4$, close to $\tfrac73$ at $K=8$). With graded quality and continuous embedding distance, the case used in practice, no coefficient closes the gap, and at $K=2$ no coefficient helps in any family.

\begin{table}[h]
  \centering
  \caption{Group objective reached by GRPO with its default coefficient and with the best coefficient on a grid, as a percentage of \tg{} (exact computation). The continuous-distance row averages three random embedding draws.}
  \label{tab:screen}
  \small
  \begin{tabular}{@{}lccc@{}}
    \toprule
    Reward family & $K=2$ & $K=4$ & $K=8$ \\
    \midrule
    Quality $\times$ same-class indicator (DARLING, binary kernel) & 0 / 0 & 88 / 100 & 93 / 99.8 \\
    Coverage with shared credit & 73 / 73 & 91 / 99 & 91 / 99.8 \\
    Graded quality $\times$ continuous cosine distance & 0.3 / 11 & 61 / 89 & 70 / 96 \\
    \bottomrule
  \end{tabular}
\end{table}

\section{Standard-deviation normalization}\label{app:std}

Standard GRPO divides each advantage by the standard deviation of the group's rewards. At $K=2$ this keeps only the sign of $r_1-r_2$, in which a symmetric pairwise term cancels, so normalization inherits GRPO's blindness exactly, sample by sample. For larger $K$ it adds a second distortion that depends on $K$ non-monotonically and has no closed form, so we compute its fixed point numerically and test it on the 9B model. \Cref{tab:std} shows the fixed point for the diversity bonus of \cref{prop:fixed}(a), at which two answers keep positive mass. Normalization moves the fixed point by up to $0.19$, and the trained 9B policies land within $0.01$ of the numerical prediction at $K=4$ and $K=8$. On GlobalOpinionQA, a trained model again, GRPO with normalization is indistinguishable from GRPO at $K=2$ ($0.508$ against $0.491$) and slightly worse at $K=4$ ($0.207$ against $0.181$, \cref{tab:opinion}). In the generality study (\cref{tab:generality}) it changes the GRPO column by at most $9$ points and never reaches \tg{} where GRPO falls short of it. We therefore use unnormalized advantages for every estimator. \Cref{app:ppo} treats the remaining components of practical GRPO, clipping and several epochs per batch, including clipping combined with normalization.

\begin{table}[h]
  \centering
  \caption{Mass on the best answer at the fixed point of the diversity bonus ($\lambda=1$): closed form, numerical ascent using exact expectations, and the 9B sandbox.}
  \label{tab:std}
  \small
  \begin{tabular}{@{}lcccc@{}}
    \toprule
    & $K=3$ & $K=4$ & $K=6$ & $K=8$ \\
    \midrule
    GRPO, closed form & 1.000 & 0.875 & 0.813 & 0.792 \\
    GRPO (std), numerical & 1.000 & 0.686 & 0.814 & 0.713 \\
    GRPO (std), Qwen3.5-9B & -- & 0.695 & -- & 0.717 \\
    \bottomrule
  \end{tabular}
\end{table}

\section{Clipping and several epochs per batch}\label{app:ppo}

PPO-style implementations of GRPO draw a batch $B$ from $\pi_{\mathrm{old}}$ and take $\mu$ gradient steps on the clipped surrogate
\begin{equation*}
  L(\theta)=\tfrac{1}{|B|}\textstyle\sum_{i\in B}\min\big(\rho_iA_i,\ \mathrm{clip}(\rho_i,1-\epsilon,1+\epsilon)\,A_i\big),\qquad \rho_i=\pth(y_i)/\pi_{\mathrm{old}}(y_i).
\end{equation*}
\begin{proposition}\label{prop:ppo}
For any advantage, any $\epsilon>0$ and any number of epochs $\mu$, a policy is left unchanged by the expected clipped update with small enough steps if and only if the plain expected update $\E\big[\sum_kA_k\sth(y_k)\big]$ vanishes there.
\end{proposition}
\begin{proof}
At $\theta=\theta_{\mathrm{old}}$ every ratio equals one, no sample is clipped, and $\nabla_\theta\E[L]=\tfrac1K\E\big[\sum_kA_k\sth(y_k)\big]$. If this vanishes, the first step leaves $\theta$ unchanged, and so does every later one. Otherwise each small enough step strictly increases $\E[L]$, so after $\mu$ steps the policy differs from $\pi_{\mathrm{old}}$.
\end{proof}

The fixed points of \cref{prop:fixed} and the numerical fixed points of normalized GRPO are therefore fixed points of the clipped update as well. Two effects remain open, the transient in which clipping is active and the reuse of a finite batch, whose later epochs fit that batch's noise. We measure both on the answer space of the sandbox (\cref{app:sandbox}) with a softmax policy, for all 14 sandbox configurations and for distribution matching (\cref{prop:fixed}b) with target $t=(0.45,0.30,0.15,0.10)$, using $\epsilon=0.2$ throughout. In the \emph{exact} version the expectation over groups is computed by enumeration, and each of 20{,}000 iterations takes $\mu=4$ steps of size $10/(1+n/500)$ at iteration $n$. In the \emph{sampled} version each iteration draws 256 rollouts and takes $\mu\in\{1,4,8\}$ Adam steps with learning rate $0.05$, over 3{,}000 iterations and 3 seeds, and the policy is averaged over the last 1{,}000 iterations. One epoch is the plain update and measures sampling noise alone.

In the exact version clipping is active on $54$--$100\%$ of samples early in training, yet every run converges to the fixed point of the plain update (\cref{tab:ppo}). The error is below $10^{-8}$ in $\ell_1$ where every answer keeps positive mass, and below $10^{-3}$ where some answers die out, since their mass vanishes only slowly under a decaying step. Clipping combined with normalization lands on the numerical fixed point of normalized GRPO in the same way. In the sampled version one epoch stays within $0.008$ of the fixed point. Reusing each batch adds a small shift toward concentration that affects every estimator, with largest error $0.025$ at $\mu=4$ and $0.074$ at $\mu=8$, while up to $2.9\%$ of samples are clipped. The shift is largest for REINFORCE and for \tg{} at $K=2$, and it stays an order of magnitude below the distance between the fixed points of GRPO and \tg{} at the same $K$, which ranges from $0.50$ to $1.10$. GRPO at $K=2$ ends on a single answer in every configuration, with or without clipping.

\begin{table}[h]
  \centering
  \caption{Clipped surrogate ($\epsilon=0.2$) against the fixed point of the plain update. The exact columns give the $\ell_1$ error at convergence and the largest share of clipped samples during training (\%). The sampled columns give the $\ell_1$ error of the time-averaged policy for $\mu$ epochs per batch and the share of clipped samples at $\mu=8$ (\%). $^\ast$Numerical fixed point of the plain update.}
  \label{tab:ppo}
  \small
  \setlength{\tabcolsep}{3.5pt}
  \begin{tabular}{@{}llcclcccccc@{}}
    \toprule
    & & & & & \multicolumn{2}{c}{Exact, $\mu=4$} & \multicolumn{4}{c}{Sampled} \\
    \cmidrule(lr){6-7}\cmidrule(l){8-11}
    Reward & Estimator & $K$ & $\lambda$ & Fixed point & $\ell_1$ & Clipped & $\mu=1$ & $\mu=4$ & $\mu=8$ & Clipped \\
    \midrule
    Diversity bonus & GRPO & 2 & 0 & (1, 0, 0, 0) & 0.0001 & 92 & 0.000 & 0.000 & 0.000 & 0.0 \\
 & GRPO & 2 & 1 & (1, 0, 0, 0) & 0.0001 & 92 & 0.000 & 0.000 & 0.000 & 0.0 \\
 & GRPO & 2 & 5 & (1, 0, 0, 0) & 0.0001 & 92 & 0.000 & 0.000 & 0.000 & 0.0 \\
 & GRPO & 4 & 1 & (.875, .125, 0, 0) & 0.0002 & 80 & 0.005 & 0.013 & 0.012 & 2.4 \\
 & GRPO & 4 & 3 & (.537, .287, .138, .037) & 0.0000 & 79 & 0.005 & 0.008 & 0.011 & 2.5 \\
 & GRPO & 8 & 1 & (.792, .208, 0, 0) & 0.0002 & 87 & 0.002 & 0.008 & 0.011 & 1.9 \\
 & GRPO (std) & 4 & 1 & (.687, .313, 0, 0)$^\ast$ & 0.0000 & 85 & 0.001 & 0.004 & 0.012 & 2.1 \\
 & GRPO (std) & 8 & 1 & (.713, .287, 0, 0)$^\ast$ & 0.0000 & 100 & 0.001 & 0.002 & 0.009 & 2.0 \\
 & REINFORCE & 2 & 1 & (.750, .250, 0, 0) & 0.0009 & 54 & 0.003 & 0.011 & 0.063 & 0.7 \\
 & REINFORCE & 4 & 1 & (.750, .250, 0, 0) & 0.0009 & 54 & 0.002 & 0.012 & 0.061 & 0.7 \\
 & REINFORCE & 8 & 1 & (.750, .250, 0, 0) & 0.0009 & 54 & 0.002 & 0.013 & 0.054 & 0.7 \\
 & \tg{} & 2 & 1 & (.537, .287, .138, .037) & 0.0000 & 64 & 0.005 & 0.025 & 0.074 & 1.8 \\
 & \tg{} & 4 & 1 & (.537, .287, .138, .037) & 0.0000 & 85 & 0.005 & 0.008 & 0.013 & 2.4 \\
 & \tg{} & 8 & 1 & (.537, .287, .138, .037) & 0.0000 & 86 & 0.003 & 0.005 & 0.007 & 2.6 \\
\addlinespace[2pt]
Distribution matching & GRPO & 2 & -- & (1, 0, 0, 0) & 0.0001 & 94 & 0.000 & 0.000 & 0.000 & 0.0 \\
 & GRPO & 4 & -- & (.725, .275, 0, 0) & 0.0002 & 80 & 0.001 & 0.009 & 0.016 & 2.2 \\
 & GRPO & 8 & -- & (.675, .325, 0, 0) & 0.0008 & 87 & 0.003 & 0.006 & 0.014 & 2.2 \\
 & REINFORCE & 4 & -- & (.633, .333, .033, 0) & 0.0004 & 75 & 0.001 & 0.006 & 0.028 & 1.3 \\
 & \tg{} & 2 & -- & (.450, .300, .150, .100) & 0.0000 & 90 & 0.008 & 0.009 & 0.011 & 2.5 \\
 & \tg{} & 4 & -- & (.450, .300, .150, .100) & 0.0000 & 79 & 0.002 & 0.005 & 0.008 & 2.6 \\
 & \tg{} & 8 & -- & (.450, .300, .150, .100) & 0.0000 & 77 & 0.003 & 0.003 & 0.003 & 2.9 \\
 
    \bottomrule
  \end{tabular}
\end{table}

\section{What the identity says about published methods}\label{app:methods}

\Cref{tab:methods} classifies the group-dependent rewards of recent methods by their interaction structure and states what \cref{cor:ladder,prop:pairwise} imply when they are optimized with GRPO. Rewards are taken from the original papers.

\begin{table}[h]
  \centering
  \caption{Group-dependent rewards in recent work and the consequence of optimizing them with group-relative baselines. \emph{Evaluated by} is the metric each paper reports for the property its reward targets.}
  \label{tab:methods}
  \small
  \setlength{\tabcolsep}{4pt}
  \begin{tabular}{@{}>{\raggedright\arraybackslash}p{2.7cm}>{\raggedright\arraybackslash}p{3.5cm}c>{\raggedright\arraybackslash}p{2.3cm}>{\raggedright\arraybackslash}p{2.7cm}>{\raggedright\arraybackslash}p{3.6cm}@{}}
    \toprule
    Method & Group-dependent term & $K$ & Structure & Evaluated by & Under GRPO \\
    \midrule
    DARLING \citep{li2025jointly} & quality $\times$ mean semantic distinctness & 8 & asymmetric pairwise & quality and novelty, pass@$k$ on math & not repairable by retuning (\cref{prop:retune}) \\
    Song et al.\ \citep{song2025outcome} & $-$ mean indicator of equal answers & 8, 16 & symmetric pairwise & pass@$k$ & keeps $\tfrac37$, $\tfrac{7}{15}$ of the weight \\
    GAPO \citep{anschel2025gapo} & penalty on the answer's frequency in the group & 32 & symmetric pairwise & diversity and coverage of valid answers & keeps $\tfrac{15}{31}$ of the weight \\
    DSDR \citep{dsdr2026} & embedding distance and formula uniqueness, gated by correctness & 8 & pairwise, gated by own correctness & pass@$k$ & diversity term attenuated \\
    SWYS \citep{swys2026} & collision with groupmates & 64 & symmetric pairwise & distance to the target distribution & fixed by their witness estimator, centred variant biased (\cref{app:witness}) \\
    Pref-GRPO \citep{wang2025prefgrpo} & win rate within the group & -- & constant-sum & per-image quality & amplified by $\tfrac{K}{K-1}$, harmless alone \\
    pass@$k$ credited to all & best correctness in the group & -- & shared & pass@$k$ & invisible to GRPO \\
    PKPO, L2O \citep{walder2025pkpo,takashiro2026l2o} & max@$k$ & -- & shared & pass@$k$, max@$k$ & their estimators are instances of \tg{} \\
    SetPO \citep{setpo2026} & leave-one-out marginal diversity & -- & marginal contribution & pass@$k$ & equals \tg{} \\
    Poly-EPO \citep{orney2026polyepo} & correctness $\times$ number of strategy clusters & 8 & set-level & pass@$k$ & own unbiased estimator, needs $N>n$ \\
    TTRL \citep{zuo2025ttrl} & agreement with the majority answer & -- & externality anti-aligned & per-sample accuracy & only the step size changes \\
    MAGRPO \citep{liu2025magrpo} & shared team reward & -- & baseline from other episodes & team reward & unaffected \\
    \bottomrule
  \end{tabular}
\end{table}

\section{Sandbox with exact readout}\label{app:sandbox}

Each prompt asks the model to choose one of four labelled options and to reply with its letter. The four letters are the first tokens of the four actions, and the policy on a prompt is read off as the softmax over these tokens. The reward is $r_k=f(a_k)+\lambda\cdot\tfrac{1}{K-1}\sum_{j\neq k}\ind[a_k\neq a_j]$ with $f=(1,0.5,0.2,0)$. We train Qwen3.5-9B with LoRA on 128 prompts, 32 prompts per step, for 500 steps, and evaluate on 64 held-out prompts. \Cref{tab:sandbox} lists all runs. The largest $\ell_1$ distance between a trained policy and its closed-form fixed point is $0.018$, and held-out prompts reproduce the training prompts to within $0.004$ per option.

\begin{table}[h]
  \centering
  \caption{All sandbox runs. Policies are averaged over prompts at the end of training.}
  \label{tab:sandbox}
  \small
  \begin{tabular}{@{}lcclllc@{}}
    \toprule
    Estimator & $K$ & $\lambda$ & Trained (train prompts) & Trained (held-out) & Closed form & $\ell_1$ \\
    \midrule
    GRPO & 2 & 0 & (1, 0, 0, 0) & (1, 0, 0, 0) & (1, 0, 0, 0) & 0.000 \\
GRPO & 2 & 1 & (1, 0, 0, 0) & (1, 0, 0, 0) & (1, 0, 0, 0) & 0.000 \\
GRPO & 2 & 5 & (1, 0, 0, 0) & (1, 0, 0, 0) & (1, 0, 0, 0) & 0.000 \\
GRPO & 4 & 1 & (.878, .122, 0, 0) & (.876, .124, 0, 0) & (.875, .125, 0, 0) & 0.006 \\
GRPO & 4 & 3 & (.538, .286, .130, .046) & (.538, .286, .130, .046) & (.537, .287, .138, .037) & 0.018 \\
GRPO & 8 & 1 & (.797, .203, 0, 0) & (.795, .205, 0, 0) & (.792, .208, 0, 0) & 0.011 \\
GRPO (std) & 4 & 1 & (.695, .305, 0, 0) & (.694, .306, 0, 0) & -- & -- \\
GRPO (std) & 8 & 1 & (.717, .283, 0, 0) & (.714, .286, 0, 0) & -- & -- \\
REINFORCE & 2 & 1 & (.758, .242, 0, 0) & (.758, .242, 0, 0) & (.750, .250, 0, 0) & 0.016 \\
REINFORCE & 4 & 1 & (.750, .248, .002, 0) & (.751, .247, .002, 0) & (.750, .250, 0, 0) & 0.005 \\
REINFORCE & 8 & 1 & (.751, .248, 0, 0) & (.752, .248, 0, 0) & (.750, .250, 0, 0) & 0.003 \\
\tg{} & 2 & 1 & (.541, .285, .136, .038) & (.541, .286, .135, .038) & (.537, .287, .138, .037) & 0.008 \\
\tg{} & 4 & 1 & (.532, .288, .136, .045) & (.534, .289, .132, .045) & (.537, .287, .138, .037) & 0.015 \\
\tg{} & 8 & 1 & (.538, .287, .135, .040) & (.538, .289, .133, .039) & (.537, .287, .138, .037) & 0.005 \\
 
    \bottomrule
  \end{tabular}
\end{table}

\section{Distributional alignment details}\label{app:opinion}

\paragraph{Data.} We use the GlobalOpinionQA release of \citet{durmus2024towards}. For each (question, country) pair we drop non-substantive options (``Don't know'', ``Refused'' and similar), renormalize the rest, and keep items with at least two options and at least half of the mass on substantive options. We split by question text with 70\% of questions for training, 10\% for development and 20\% for testing, and keep at most three countries per question in the development and test splits. This yields 32{,}372 training items, 618 development items and 1{,}154 test items. The prompt reads ``Imagine you are a randomly selected adult living in [country], answering a public opinion survey.'', followed by the question, the lettered options, and the instruction to reply with the letter only.

\paragraph{Training and evaluation.} All runs use 256 rollouts per step, learning rate $2\cdot10^{-5}$, at most 16 new tokens, 200 steps, and checkpoints every 50 steps. Answers that do not parse to a valid letter count as a separate outcome. For each test item we draw 32 samples with counts $n_a$ and estimate the squared error without bias as $\sum_a\tfrac{n_a(n_a-1)}{n(n-1)}-2\sum_a t_a\tfrac{n_a}{n}+\|t\|^2$. Paired confidence intervals resample questions with all their countries.

\paragraph{All checkpoints.} \Cref{tab:opinion-ckpt} reports checkpoint metrics averaged over seeds for each estimator and group size. Poly-EPO is listed under the number of samples it draws per question. \tg{} has lower error than GRPO at every checkpoint and in every seed. Differences between seeds ($\le0.042$) are an order of magnitude smaller than differences between GRPO and \tg{} ($\approx0.4$ at $K=2$). The general and symmetric forms of \tg{} have the same expected update, and the symmetric form ends lower ($0.074$ against $0.086$ at $K=4$) with a baseline from other questions that preserves the expected update.

\begin{table}[h]
  \centering
  \caption{Test error $\|\pi-t\|^2$ at each checkpoint, and TV distance and collapse rate after 200 steps, averaged over seeds for each estimator and $K$. The base model has error $0.197$, TV $0.330$ and collapse rate $16.6\%$.}
  \label{tab:opinion-ckpt}
  \small
  \begin{tabular}{@{}clcccccc@{}}
    \toprule
    $K$ & Estimator & 50 & 100 & 150 & 200 & TV & Coll.\ (\%) \\
    \midrule
    2 & GRPO & 0.528 & 0.514 & 0.505 & 0.491 & 0.515 & 96.5 \\
\addlinespace[1pt]
2 & GRPO (std) & 0.539 & 0.534 & 0.510 & 0.508 & 0.523 & 96.1 \\
\addlinespace[1pt]
2 & REINFORCE & 0.130 & 0.158 & 0.131 & 0.131 & 0.268 & 13.3 \\
\addlinespace[1pt]
2 & \tg{} & 0.097 & 0.087 & 0.082 & 0.081 & 0.239 & 0.5 \\
\addlinespace[1pt]
4 & GRPO & 0.194 & 0.213 & 0.196 & 0.181 & 0.311 & 28.8 \\
\addlinespace[1pt]
4 & GRPO (std) & 0.204 & 0.224 & 0.200 & 0.207 & 0.329 & 37.8 \\
\addlinespace[1pt]
4 & Centred witness & 0.111 & 0.113 & 0.107 & 0.096 & 0.239 & 7.1 \\
\addlinespace[1pt]
4 & Poly-EPO (pairs) & 0.100 & 0.087 & 0.086 & 0.078 & 0.230 & 0.6 \\
\addlinespace[1pt]
4 & \tg{} (general form) & 0.099 & 0.091 & 0.084 & 0.086 & 0.238 & 2.9 \\
\addlinespace[1pt]
4 & \tg{} (symmetric form) & 0.093 & 0.085 & 0.077 & 0.074 & 0.228 & 0.2 \\
\addlinespace[1pt]
8 & GRPO & 0.187 & 0.159 & 0.153 & 0.179 & 0.308 & 31.8 \\
\addlinespace[1pt]
8 & Centred witness & 0.102 & 0.090 & 0.085 & 0.073 & 0.217 & 1.8 \\
\addlinespace[1pt]
8 & Poly-EPO (pairs) & 0.098 & 0.095 & 0.089 & 0.074 & 0.225 & 0.8 \\
\addlinespace[1pt]
8 & \tg{} (symmetric form) & 0.103 & 0.093 & 0.090 & 0.077 & 0.225 & 1.6 \\
 
    \bottomrule
  \end{tabular}
\end{table}

\paragraph{Equal distance from the base model.} To check that \tg{} does not win by moving more or less, we measure how far each checkpoint's answer distribution has moved from the base model's (an unbiased estimate of $\|\pi-\pi_0\|^2$, averaged over test items) and interpolate the error of each run at equal distance. \Cref{tab:opinion-dist} reports the difference at half and at all of the smaller of the two final distances. GRPO moves furthest from the base model at $K=2$ and its error increases while it does so, so it is moving in the wrong direction.

\begin{table}[h]
  \centering
  \caption{Difference in test error at equal distance from the base model (95\% CIs over questions).}
  \label{tab:opinion-dist}
  \small
  \begin{tabular}{@{}lcc@{}}
    \toprule
    Comparison & half distance & full distance \\
    \midrule
    \tg{} $-$ GRPO, $K=2$, seed 0 & $-0.115$ {\scriptsize$[-0.122,-0.107]$} & $-0.207$ {\scriptsize$[-0.219,-0.195]$} \\
    \tg{} $-$ GRPO, $K=2$, seed 1 & $-0.110$ {\scriptsize$[-0.117,-0.104]$} & $-0.217$ {\scriptsize$[-0.229,-0.204]$} \\
    \tg{} (general form) $-$ GRPO, $K=4$ & $-0.076$ {\scriptsize$[-0.081,-0.070]$} & $-0.121$ {\scriptsize$[-0.131,-0.111]$} \\
    \tg{} $-$ GRPO, $K=8$ & $-0.071$ {\scriptsize$[-0.078,-0.064]$} & $-0.088$ {\scriptsize$[-0.100,-0.077]$} \\
    \bottomrule
  \end{tabular}
\end{table}

\section{Open-ended generation details}\label{app:novelty}

\paragraph{Training data and reward.} We select 4{,}000 training prompts from WildChat \citep{zhao2024wildchat} following the NoveltyBench recipe and remove every evaluation prompt. The quality score $q$ is the raw reward of Skywork-Reward-V2-Llama-3.1-8B mapped through $\sigma((x-12.7)/8.8)$, with location and scale calibrated on base-model outputs. The distance $d$ is the cosine distance between e5-large-v2 embeddings, linearly rescaled to $[0,1]$, and $D_k$ averages it over the groupmates. Generations use temperature 1 and at most 512 new tokens.

\paragraph{Evaluation.} We evaluate on 400 NoveltyBench prompts, all 100 curated prompts and the first 300 WildChat prompts, with 10 samples per prompt at temperature 1 and at most 512 tokens. Equivalence classes come from the official DeBERTa classifier, and utility uses Skywork-Reward-Gemma-2-27B-v0.2 with the official thresholds and patience $0.8$ \citep{zhang2025noveltybench}. For the LLM judge we follow the NoveltyBench v1.1 protocol with its official grouping and scoring instructions, using Claude Sonnet 5.5 as the judge. We regenerate with 2{,}048 tokens, sample 100 prompts in the same curated-to-WildChat proportion, and present samples anonymized and shuffled across models.

\begin{table}[h]
  \centering
  \caption{NoveltyBench under the LLM judge (NoveltyBench v1.1 protocol, 100 prompts). $\Delta$ is \tg{} minus GRPO with 95\% CIs. Bold is the best trained model on set metrics.}
  \label{tab:judge}
  \small
  \begin{tabular}{@{}lcccccc@{}}
    \toprule
    & \multicolumn{3}{c}{$K=2$} & \multicolumn{3}{c}{$K=8$} \\
    \cmidrule(lr){2-4}\cmidrule(l){5-7}
    & Utility $\uparrow$ & Distinct $\uparrow$ & 1st sample & Utility $\uparrow$ & Distinct $\uparrow$ & 1st sample \\
    \midrule
    Base model & 2.27 & 4.09 & 5.11 & 2.27 & 4.09 & 5.11 \\
    GRPO ($\lambda{=}2$) & 2.10 & 2.27 & 5.82 & 2.35 & 2.81 & 5.77 \\
    \textbf{\tg{}} ($\lambda{=}2$) & \textbf{2.71} & \textbf{4.36} & 5.50 & \textbf{2.61} & \textbf{4.15} & 5.55 \\
    $\Delta$, same step & $+0.61$ {\scriptsize$[0.36,0.85]$} & $+2.09$ {\scriptsize$[1.67,2.52]$} & & $+0.26$ {\scriptsize$[0.07,0.45]$} & $+1.34$ {\scriptsize$[1.05,1.63]$} & \\
    $\Delta$, matched KL & $+0.41$ {\scriptsize$[0.24,0.58]$} & $+1.41$ {\scriptsize$[1.13,1.70]$} & & $+0.17$ {\scriptsize$[0.03,0.31]$} & $+0.79$ {\scriptsize$[0.49,1.08]$} & \\
    \bottomrule
  \end{tabular}
\end{table}

\paragraph{The two-sample baseline.} \Cref{tab:k2var} compares unbiased choices of $b_k$ for \tg{} at $K=2$ on real training batches. All are unbiased because none depends on $y_k$. Predicting the missing response's quality from the teammate's, with the within-prompt correlation of quality ($0.74$) estimated on earlier steps, gives the lowest second moment. A learned prompt-level regressor reduces it by only a further $7\%$, so a critic is not worth its cost. GRPO's smaller second moment does not indicate a better estimator. Its advantage \eqref{eq:k2adv} keeps only the quality difference within the pair and cancels the shared diversity term, so its magnitude is smaller exactly by the signal that \tg{} carries.

\begin{table}[h]
  \centering
  \caption{Second moment of the \tg{} advantage at $K=2$ for different baselines ($\lambda=1$, real batches).}
  \label{tab:k2var}
  \small
  \begin{tabular}{@{}lc@{}}
    \toprule
    Baseline $b_k$ & $\E[A^2]$ \\
    \midrule
    Missing response's diversity set to zero & 0.145 \\
    Batch means of quality and distance & 0.078 \\
    Batch-mean distance, quality shrunk toward the teammate (used) & 0.056 \\
    GRPO, rescaled to the same scale (reference) & 0.035 \\
    \bottomrule
  \end{tabular}
\end{table}

\paragraph{Diversity coefficient.} \Cref{tab:lambda} reports the $\lambda$ sweep behind \cref{fig:novelty}d. GRPO collapses more as $\lambda$ grows. \tg{} becomes more diverse as $\lambda$ grows. At $\lambda=0.25$ the two estimators end close together, as the theory requires, because with so small a diversity weight the optimum of $J$ itself is close to the quality-only policy that GRPO converges to. The curves for $\lambda=0.25$ and $\lambda=4$ average fewer runs than the others, which is why they fluctuate more in \cref{fig:novelty}d.

\begin{table}[h]
  \centering
  \caption{Diversity coefficient sweep with the benchmark scorer at $K=2$, step 150. Training diversity is the mean pairwise embedding distance over the last 25 training updates.}
  \label{tab:lambda}
  \small
  \begin{tabular}{@{}lccccc@{}}
    \toprule
    Estimator & $\lambda$ & Training diversity & Utility & Distinct & First sample \\
    \midrule
    GRPO & 0.25 & 0.193 & 3.80 & 3.10 & 9.12 \\
GRPO & 1 & 0.186 & 3.83 & 3.19 & 9.07 \\
GRPO & 2 & 0.147 & 3.44 & 2.65 & 9.04 \\
\tg{} & 0.25 & 0.166 & 3.84 & 3.18 & 9.09 \\
\tg{} & 1 & 0.231 & 4.25 & 4.01 & 8.72 \\
\tg{} & 2 & 0.249 & 4.61 & 4.56 & 8.51 \\
 
    \bottomrule
  \end{tabular}
\end{table}

\paragraph{Examples.} \Cref{tab:examples} shows what the three models answer on three prompts from the LLM-judge evaluation. On each, GRPO settles on one answer and repeats it almost verbatim, picking the same ``random'' prime in seven of ten samples and the same icebreaker in eight. \tg{} spreads its ten samples over many good answers, and its first sample is rated at least as high as GRPO's on all three prompts ($8$ against $8$, $8$ against $7$ and $5$ against $4$). On the filly prompt it also leads with one recommended name, as the prompt asks, where the base model and GRPO return lists.

\begin{table}[h]
  \centering
  \caption{All ten samples for three NoveltyBench prompts from the LLM-judge evaluation ($K=2$, $\lambda=2$, step 150), summarized by the answer each sample gives, with the judge's number of distinct answers and set utility. $\times n$ marks an answer given in $n$ samples.}
  \label{tab:examples}
  \small
  \setlength{\tabcolsep}{4pt}
  \renewcommand{\arraystretch}{1.15}
  \begin{tabular}{@{}>{\raggedright\arraybackslash}p{2.9cm}>{\raggedright\arraybackslash}p{4.2cm}>{\raggedright\arraybackslash}p{4.4cm}>{\raggedright\arraybackslash}p{4.6cm}@{}}
    \toprule
    Prompt & Base model & GRPO & \tg{} \\
    \midrule
    ``Randomly pick a prime number between 1 and 50.''
    & 29 $\times4$, 17, 23, 37, 43, 47, and one sample listing four primes.\newline \emph{distinct 7, utility 6.42}
    & 23 $\times7$, 29 $\times2$, 17. All ten use the same template, and two garble the list of primes they append.\newline \emph{distinct 3, utility 3.70}
    & 29 $\times3$, 19 $\times2$, 37 $\times2$, 11, 23, 31, mostly as a bare number.\newline \emph{distinct 6, utility 6.45} \\
    \addlinespace[3pt]
    ``I need you to give me the single best ice breaking question.''
    & ``If you could instantly master any skill, what would it be?'' $\times4$, and six other questions.\newline \emph{distinct 6, utility 4.64}
    & ``What's something you're really passionate about right now?'' $\times8$, and the same question with ``looking forward to'' and ``excited about''.\newline \emph{distinct 3, utility 2.78}
    & A small win you had recently $\times2$, something you're quietly excited about this week $\times2$, a skill you would master instantly $\times2$, something you're learning this year, something you're quietly proud of, something that surprised even you, a simple joy.\newline \emph{distinct 7, utility 6.29} \\
    \addlinespace[3pt]
    ``Suggest a name for a dappled-gray filly living in the mountains.''
    & A list of names grouped by theme in every sample, six of them headed by Moss.\newline \emph{distinct 3, utility 2.27}
    & A short bulleted list in every sample, six of them headed by Mist, often followed by Fawn and Stone.\newline \emph{distinct 2, utility 1.61}
    & One recommended name with a rationale per sample, Cloudstrider $\times2$, Ashen Peak, Mistweave, Mistynose, Hemlock Storm, Summit Sprout, Mistborn, Moss Run, Frostbit. Half add a few alternatives.\newline \emph{distinct 9, utility 5.69} \\
    \bottomrule
  \end{tabular}
\end{table}

The benchmark scorer shows the same pattern. For the prompt ``What is the single best first phone for a teenager?'', which asks for exactly one recommendation, the base model recommends the Google Pixel 8a in six of ten samples and names four phones in total. After training, GRPO recommends the Pixel 7a in seven of ten samples and still names only three phones. \tg{} names six different phones across the ten samples (iPhone 13, 15 and 16, Pixel 8a, Galaxy A54 and A55), each with its own justification.

\end{document}